\documentclass[11pt]{article}

\usepackage[final]{acl}
\usepackage{times}
\usepackage{latexsym}
\usepackage{amsmath}
\usepackage{fontawesome5}
\usepackage{amssymb}    
\usepackage{algorithm}   
\usepackage{algorithmic} 
\usepackage{multirow}
\usepackage{amsthm}

\usepackage{mathtools}
\usepackage{multirow}
\usepackage[table,xcdraw]{xcolor}
\usepackage[normalem]{ulem}
\useunder{\uline}{\ul}{}
\usepackage{booktabs}
\usepackage{array}
\usepackage{colortbl}
\usepackage{xcolor}

\usepackage{subcaption}

\newcommand{\algComment}[1]{ \textcolor{green!60!black}{// #1}}

\usepackage{hyperref} 
\usepackage{graphicx} 

\newcommand{\hfemoji}{\raisebox{-0.1em}{\includegraphics[height=0.9em]{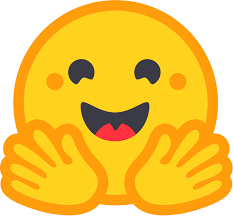}}}
\usepackage[T1]{fontenc}

\usepackage[utf8]{inputenc}

\usepackage{microtype}

\usepackage{inconsolata}

\usepackage{graphicx}
\usepackage{amsmath}
\usepackage{xcolor}
\usepackage{booktabs}
\usepackage{graphicx}

\definecolor{posgreen}{RGB}{34,139,34} 
\definecolor{negorange}{RGB}{255,69,0} 

\definecolor{refcolor}{HTML}{B3A369}
\hypersetup{
    colorlinks=true,
    linkcolor=refcolor,
    citecolor=refcolor,
    filecolor=magenta,      
    urlcolor=refcolor,
    }

\newcommand{\res}[2]{%
  #1 \ifdim #2pt>0pt%
    \textcolor{posgreen}{\scriptsize{(+#2)}}%
  \else\ifdim #2pt<0pt%
    \textcolor{negorange}{\scriptsize{(#2)}}%
  \else%
    \scriptsize{(0.00)}%
  \fi\fi%
}

\title{Behavior Knowledge Merge in Reinforced Agentic Models}

\author{
  Xiangchi Yuan$^{1}$,
  Dachuan Shi$^{1}$,
  Chunhui Zhang$^{2}$,
  Zheyuan Liu$^{3}$, \\
  \textbf{Shenglong Yao$^{1}$,
  Soroush Vosoughi$^{2}$,
  Wenke Lee$^{1}$} \\
  $^{1}$Georgia Institute of Technology,
  $^{2}$Dartmouth College,
  $^{3}$University of Notre Dame \\
  \texttt{xyuan300@gatech.edu} \\ 
  \vspace{0.8em} 
  \large 
  \href{https://xiangchi-yuan.github.io/ram-project/}{\color{refcolor}\faGlobe\ Project Page} 
  \hspace{2em} 
  \href{https://github.com/xiangchi-yuan/mrl}{\color{refcolor}\faGithub\ Code}
}

\begin{document}
\maketitle

\begin{abstract}
Reinforcement learning (RL) is central to post-training, particularly for agentic models that require specialized reasoning behaviors. In this setting, model merging offers a practical mechanism for integrating multiple RL-trained agents from different tasks into a single generalist model. However, existing merging methods are designed for supervised fine-tuning (SFT), and they are suboptimal to preserve task-specific capabilities on RL-trained agentic models.
The root is a task-vector mismatch between RL and SFT: on-policy RL induces task vectors that are highly sparse and heterogeneous, whereas SFT-style merging implicitly assumes dense and globally comparable task vectors. When standard global averaging is applied under this mismatch, RL's non-overlapping task vectors that encode critical task-specific behaviors are reduced and parameter updates are diluted.
To address this issue, we propose \textbf{R}einforced \textbf{A}gent \textbf{M}erging (\textbf{RAM}), a distribution-aware merging method explicitly designed for RL-trained agentic models. RAM disentangles shared and task-specific unique parameter updates, averaging shared components while selectively preserving and rescaling unique ones to counteract parameter update dilution. 
Experiments across multiple agent domains and model architectures demonstrate that RAM not only surpasses merging baselines, but also unlocks synergistic potential among agents to achieve performance superior to that of specialized agents in their domains.

\end{abstract}

\section{Introduction}

Post-training has become a cornerstone for aligning large language models (LLMs) to diverse domains~\cite{wei2021finetuned,liu2025modality,ouyang2022training,jaech2024openai,tan2024democratizing}. While specializing a model for a single task is effective, real-world applications typically require a single model to possess multi-task capabilities. Traditionally, these capabilities are obtained by mixing offline datasets and performing joint training. However, training such a generalist model from scratch is computationally expensive, and maintaining separate checkpoints for each task~\cite{chen2024agentverse,jin2025two} is storage-inefficient. Consequently, model merging, which combines multiple task-specific models fine-tuned from the same base model into a unified model, has emerged as a widely adopted solution~\citep{ilharcoediting,liu2024towards,matena2022merging,yu2024language}. It offers multiple advantages, including data privacy preservation, the elimination of additional training costs, and minimal sacrificed performance.

\begin{figure}[t]
    \centering
    \includegraphics[width=0.95\linewidth]{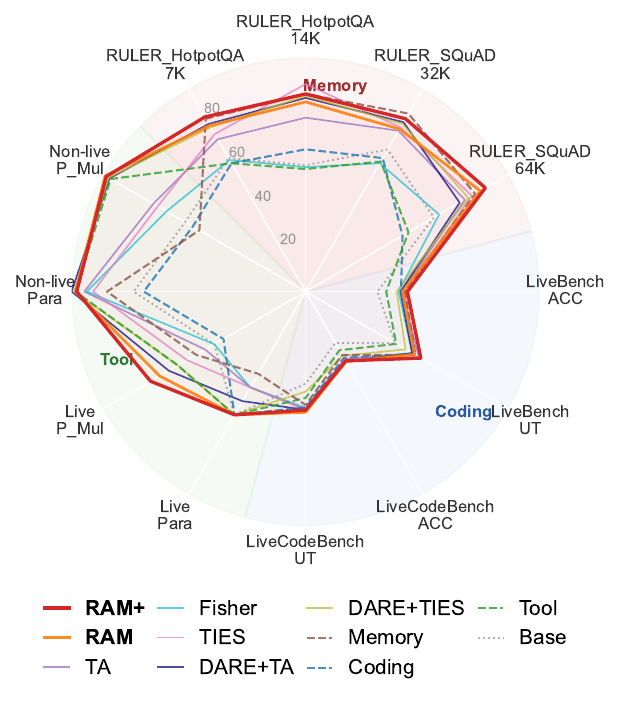}
    \vspace{-0.3cm}
    \caption{Performance comparison of RAM/RAM+ and baselines on 12 tasks across 3 agent domains. Our method achieves the best average performance and secures SOTA results on 9 out of 12 tasks, surpassing even the original specialized agents (Coding, Memory, Tool).}
    \vspace{-0.3cm}
    \label{fig:intro}
\end{figure}

Recently, post-training has shifted from supervised fine-tuning (SFT) to reinforcement learning (RL)~\citep{team2025kimi,guo2025deepseek}, particularly for agentic models with strong reasoning capabilities. This shift fundamentally changes the role of model merging. In the SFT scenarios, multi-task capabilities can still be easily obtained by performing joint training. In contrast, joint multi-task training under on-policy RL is impractical in real-world applications, as it requires parallel task-specific environments and reward models to ensure on-policy training. Consequently, model merging becomes a convenient solution in the RL setting.
A representative example from large-scale industrial agents is UI-TARS2~\citep{wang2025ui}, which trains specialized vertical agents in isolated environments via RL, and subsequently merges them into a unified generalist agent. This paradigm reflects a practical compromise: specialization through RL, followed by post-hoc integration through model merging.

However, directly applying existing model merging methods to RL-trained agents like UI-TARS2 leads to performance degradation. Most prior approaches, including Task Arithmetic~\citep{ilharcoediting}, TIES-Merging~\citep{yadav2023resolving}, and DARE~\citep{yu2024language}, are developed under the assumption of SFT parameter updates (task vectors) and are therefore mismatched to the RL setting. Unlike SFT, which typically induces dense and redundant parameter updates~\citep{chu2025sft,shenfeld2025rls}, on-policy RL produces highly sparse and often disjoint task vector distributions, shaped by task-specific reward signals and RL objectives that target narrow behaviors. When such sparse updates are globally averaged during merging, task-specific unique parameter updates are divided by the number of models, resulting in signal dilution, which degrades task-specific behavior knowledge.

To address this mismatch, we propose \textbf{R}einforced \textbf{A}gent \textbf{M}erging (\textbf{RAM}), a distribution-aware merging method designed specifically for RL-trained agents. RAM explicitly disentangles shared and task-specific unique regions of task vectors obtained via RL processes, averaging shared regions to preserve common capabilities while selectively preserving and rescaling unique regions to prevent signal dilution. By maintaining RL task vectors during merging, RAM enables specialized behavior knowledge from each model to coexist within the merged unified model.
Our contributions for reinforced agent merging are: 
\begin{itemize} 
\item 
We identify the mismatch between merging RL-trained agentic models 
and the existing merging method for SFT-trained models, which is signal dilution in the heterogeneous distribution of sparse parameter updates.  
\item 
We propose a distribution-aware merging method that treats shared and task-unique parameter updates differently, averaging the former while preserving and rescaling the latter with distribution, avoiding signal dilution and compensating for performance degradation.

\item Extensive experiments demonstrate that RAM not only outperforms existing merging methods across diverse architectures and domains but also unlocks synergistic potential among agents. The unified RAM model achieves performance superior to that of individual specialized agents on their domain tasks.

\end{itemize}

\section{Related Works}

\paragraph{Post-training Agents with RL}
RL has recently emerged as a pivotal paradigm for enhancing the reasoning capabilities of Large Language Models (LLMs)~\citep{jaech2024openai, shao2024deepseekmath, team2025kimi,jin2025your,diao2026addressing}. Several general-purpose algorithms, such as PPO~\citep{schulman2017proximal}, GRPO~\citep{guo2025deepseek}, and DAPO~\citep{yu2025dapo}, have been developed to support this direction. Beyond general reasoning, RL is extensively applied to specialize agents for diverse domains. In coding, methods like CURE~\cite{wang2025cure}, SWE-RL~\cite{wei2025swe}, and GLM-4.5~\cite{zeng2025glm} have demonstrated significant success. For memory extension, MemAgent~\cite{yu2025memagent} and various cache-based approaches~\cite{shilacache} optimize long-context handling. Furthermore, RL has been instrumental in developing tool-integrated reasoning agents, such as AutoTIR~\cite{wei2025autotir} and ToolRL~\cite{qian2025toolrl}, as well as search-augmented agents~\cite{jin2025searchr,tongyidr,sun2025zerosearch} and computer-use agents~\cite{wang2025ui,ye2025mobile}.

\paragraph{Model Merging for LLMs}
Model merging has demonstrated superiority in multi-task learning by synthesizing different task-specific models into a single entity without additional training~\citep{ilharcoediting,jin2025search,matena2022merging}. Techniques such as Task Arithmetic~\citep{ilharcoediting}, TIES-Merging~\cite{yadav2023resolving}, and DARE~\citep{yu2024language} have been successfully validated for merging SFT models across multiple tasks. Beyond multi-task integration, merging has also been employed to mitigate model collapse~\cite{yuan2025superficial} and enhance reasoning efficiency~\citep{wu2025revisiting}. While recent works like UI-TARS2~\citep{wang2025ui} attempt to merge RL-trained agentic models, they rely on simple weight interpolation, which remains suboptimal for this regime. Distinguished from prior studies, our work is the first to systematically characterize the unique behaviors of task vectors induced by RL and to design a tailored merging strategy, that aligns their specific heterogeneity.

\section{Reinforced Task Vector Behaviors}
\begin{figure}[t]
    \centering
    \includegraphics[width=0.95\linewidth]{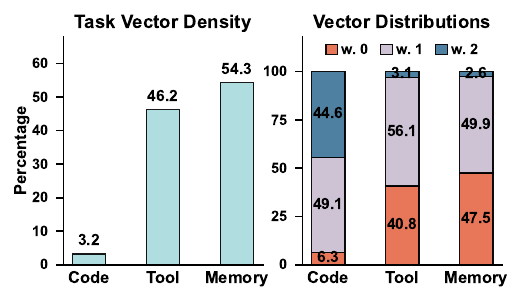}
    \caption{ \textbf{Left}: Density (1-Sparsity) of task vectors varies between agent models. \textbf{Right}: Non-zero elements distributions of task vector varies on the number of overlaps with other task vectors.}
    \label{fig:overlap-analysis}
\end{figure}

In this section, we characterize the properties of task vectors derived from reinforcement learning and explain how these properties render existing model merging methods suboptimal.

\subsection{Preliminaries and Settings}

\paragraph{Task Vectors} 
Task vector is the set of parameter updates for a specific task. Let $\theta_{\text{pre}} \in \mathbb{R}^d$ denote the parameters of a pre-trained base model. We consider $N$ different tasks, where each task $t \in \{1, \dots, N\}$ yields a fine-tuned model $\theta_t$. The task vector for task $t$ is defined as $\boldsymbol{\tau}_t = \theta_t - \theta_{\text{pre}}$. In this study, we specifically examine task vectors induced by RL fine-tuning, referring to them as \textbf{Reinforced Task Vectors}. The objective of model merging is to synthesize a single merged task vector $\boldsymbol{\tau}_{\text{merged}}$ to construct a final model $\theta_{\text{merged}} = \theta_{\text{pre}} + \boldsymbol{\tau}_{\text{merged}}$.

\paragraph{Vector Sparsity} 
We define the task vector sparsity of a model $\theta_t$ relative to the base model as $\text{sparsity}(\theta_t, \theta_{\text{pre}}) := 1 - \|\theta_t - \theta_{\text{pre}}\|_0 / d$, where $\|\cdot\|_0$ represents the number of non-zero elements and $d$ is the total parameter dimension. Following standard practice\footnote{PyTorch uses $10^{-5}$ as the default tolerance for gradient checking (refer to \href{https://pytorch.org/docs/stable/generated/torch.autograd.gradcheck.gradcheck.html}{PyTorch Documentation}).}~\cite{paszke2019pytorch,mukherjee2025reinforcement}, we consider two parameters equal (implying a zero element in the task vector) if their absolute difference is $\leq 10^{-5}$.

\paragraph{Reinforced Agentic Models}
To investigate these properties, we select three representative agentic models trained via RL to specialize in coding, tool-use, and long-context memory as follows.
\textbf{CURE}~\citep{wang2025cure}: A coding agent that co-evolves with a unit tester to enhance code generation capabilities.
\textbf{ToolRL}~\citep{qian2025toolrl}: A reasoning agent optimized for general-purpose tool selection and application.
\textbf{MemAgent}~\cite{yu2025memagent}: An agent optimized for long-context tasks, with a workflow for extended memory retention.
All reinforced agents are initialized from the same base model, Qwen2.5-7B-Instruct~\citep{qwen2}. To evaluate their specialized agentic capabilities and reinforced task vector behaviors, we employ benchmarks across the coding, tool-use, and memory domains; detailed evaluation settings are provided in Section~\ref{sec:exp:setup}.

\subsection{Heterogeneity in Reinforced Task Vectors}
\label{sec:behavior_analysis}

\begin{figure}[t]
    \centering
    \includegraphics[width=0.95\linewidth]{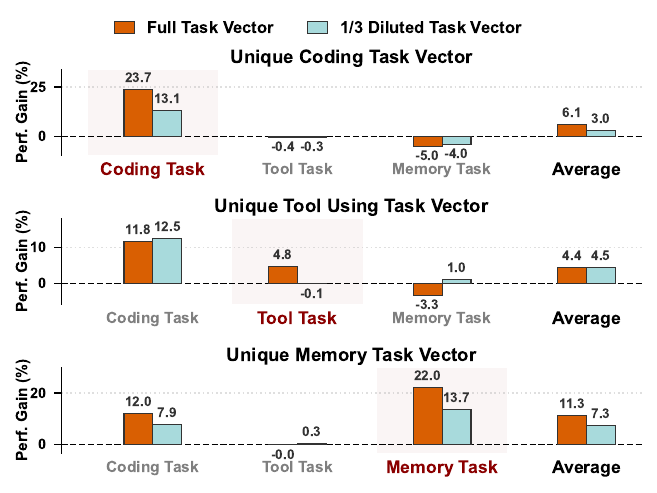}
    
    \caption{The performance gain (\%) of merging unique regions of reinforced task vectors across domains.} 
    \label{fig:unique_para}
\end{figure}

In this section, we have the two key observations by analyzing the distribution of reinforced task vectors in parameter space: 1) Reinforced task vectors are sparsely distributed in parameter space, with notable heterogeneous sparsity and distributions. 2) Heterogeneity causes shared and unique regions of reinforced task vectors. Unique regions are critical for improving corresponding agentic domain performances, and exert little negative interference on other domains' performances.

\paragraph{Heterogeneity in Sparsity and Distribution} 
Recent work~\cite{mukherjee2025reinforcement,yuan2025mitigating} highlights the inherent sparsity of RL updates: unlike SFT, which updates global parameters, RL tends to fine-tune specific sub-networks. Our analysis confirms this and further unveils heterogeneous patterns in both sparsity and spatial distribution of these updates.

First, the percentage of non-zero elements, or the sparsity levels of reinforced task vectors vary drastically. As shown in Figure~\ref{fig:overlap-analysis} (left), the coding agent exhibits extreme sparsity, modifying only 3.2\% of parameters. In contrast, agents optimized for tool use and long-context memory induce significantly denser updates, affecting 46.2\% and 54.3\% of the parameter space, respectively.

Second, these non-zero elements are distributed across disparate regions, creating different spatial overlap patterns. We categorize parameters (task vector elements) based on whether they are updated by a single agent (unique) or multiple agents (shared). Figure~\ref{fig:overlap-analysis} (right) reveals that the coding, tool-use, and memory agents concentrate different fractions of their updates in unique, non-overlapping regions (6.3\%, 40.8\%, and 47.5\% of their respective non-zero elements). Analysis in Appendix~\ref{app:add_behavior} confirms that this heterogeneity generalizes to other reinforced agents with different model architectures and specialized domains.


\begin{figure*}[t]  
    \centering
    \includegraphics[width=\textwidth]{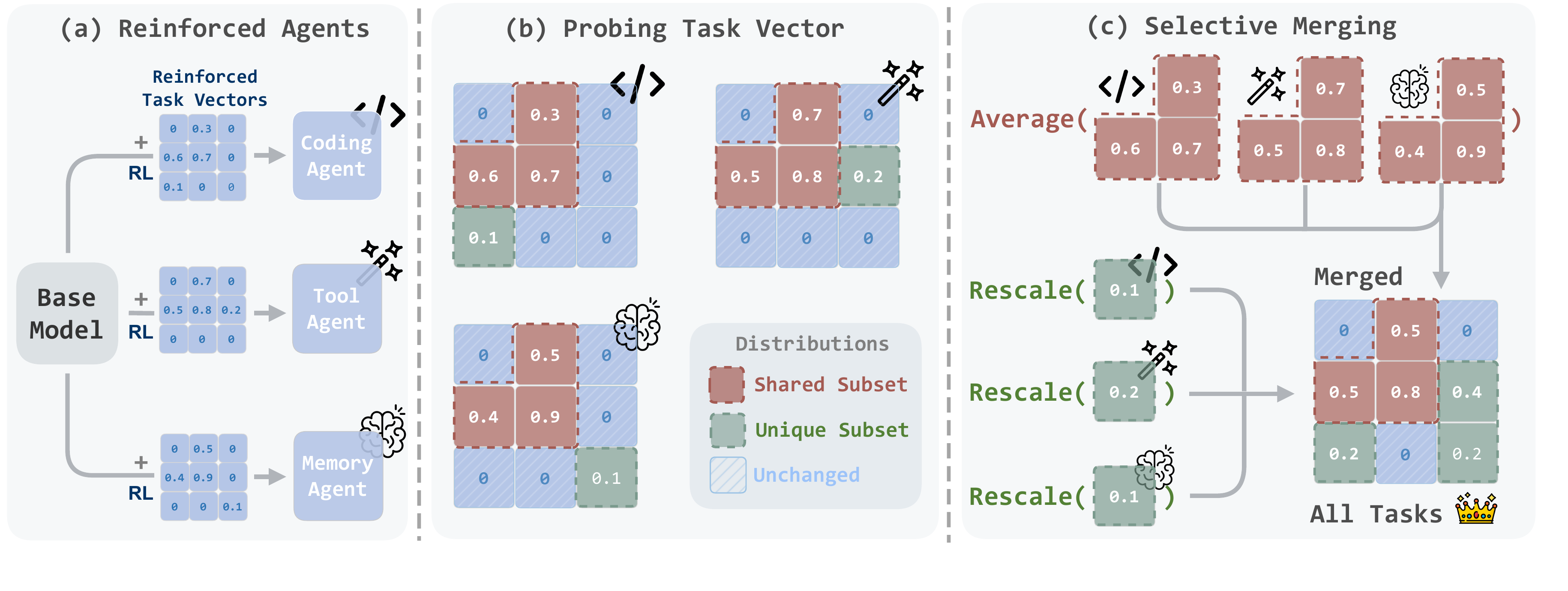} 
    \vspace{-1.2cm}
    \caption{\textbf{Method Overview.} \textbf{(a)} A base model is trained via RL to different agents, we track the distributions of obtained reinforced task vectors. \textbf{(b)} Probing the distribution of task vectors to shared, unique, unchanged sets. \textbf{(c)} Selective merging task vectors by averaging shared regions and rescaling unique regions to the base model.}
    \label{fig:method}
\end{figure*}

\paragraph{The Role of Heterogeneity}
We further investigate the impact of these heterogeneous, unique regions on both in-domain and out-of-domain agentic tasks. In our experiments, we isolate the unique component of a single task vector, merge it into the base model, and evaluate performance across all domains.
Results in Figure~\ref{fig:unique_para} demonstrate that unique regions cause almost no negative interference on out-of-domain tasks (occasionally yielding improvements) while driving significant gains on the corresponding in-domain tasks. When we intentionally dilute the magnitude of these unique vectors to $1/N (N=3)$ of their original value, in-domain performance drops sharply, verifying the positive contribution of these unique components. 

\paragraph{Signal Dilution}
Previous heterogeneity analysis uncovers the root cause of performance degradation when using previous SOTA merging methods on RL models. Existing methods typically employ element-wise averaging or variants (e.g., $\boldsymbol{\tau}_{\text{merged}} = \frac{1}{N}\sum \boldsymbol{\tau}_i$). 
While averaging is beneficial for shared regions of task vectors by balancing multi-task performance, it is detrimental to task-specific, unique regions in RL scenarios. For a unique parameter update for task $t$, averaging with $N-1$ zero-valued updates effectively scales its magnitude by $1/N$. This operation dilutes the learned signal without providing any balancing benefit, as these regions hardly interfere with out-of-domain tasks. We term this phenomenon \textit{Signal Dilution}. Figure~\ref{fig:unique_para} illustrates that this dilution (simulated with $N=3$) causes significant performance regression. The prevalence of unique task vector components in RL agents therefore necessitates a merging strategy capable of disentangling these regions to prevent signal dilution. We provide a detailed analysis of signal dilution in each existing merging strategy in Appendix~\ref{app:baseline}.



\section{Merging Reinforced Agentic Models}
\label{sec:method}
Our analysis in Section~\ref{sec:behavior_analysis} demonstrates that reinforced task vectors are inherently sparse and heterogeneous. While task vectors or parameters updated by multiple agents (shared regions) benefit from averaging to stabilize the consensus direction, parameters updated 
by a single agent (unique regions) suffer from \textit{Signal Dilution} when standard averaging is applied. 
To address this, we propose \textbf{R}einforced \textbf{A}gent \textbf{M}erging (\textbf{RAM}) illustrated in Figure~\ref{fig:method}, a method that explicitly disentangles these regions based on distribution statistics of task vectors. RAM applies selective merging strategies: it averages shared parameters to absorb unified multi-task capabilities while preserving the full magnitude of unique parameters to prevent signal dilution. Additionally, we introduce a distribution-aware rescaling mechanism to further amplify unique task capabilities.

\subsection{Probing Vector Distribution}
\label{sec:method:probing}
First, we probe the active updated parameters for each reinforced task vector $\boldsymbol{\tau}_t$. Using the threshold established in Section~\ref{sec:behavior_analysis} (e.g., $\epsilon = 10^{-5}$), we compute a binary mask $\mathbf{m}_t \in \{0, 1\}^d$ for task $t$:
\begin{equation}
\label{eq:para_stat}
    m_{t, i} = \mathbb{I}(|\tau_{t, i}| > \epsilon),
\end{equation}
where $i$ indexes the parameter dimensions and $\mathbb{I}(\cdot)$ is the indicator function. 
We then define the overlap count vector $\mathbf{c} = \sum_{t=1}^N \mathbf{m}_t$, where $c_i \in \{0, \dots, N\}$ represents the number of agents that actively update the $i$-th parameter.

For any given task $t$, the set of total updated parameters is partitioned into two disjoint subsets: the \textbf{Shared} subset (where $c_i \ge 2$) and the \textbf{Unique} subset (where $c_i = 1$). To quantify the structural distribution of the agent's updates, we define the \textit{Overlap-Unique Ratio} $\rho_t$:
\begin{equation}
\label{eq:overlap-unique}
    \rho_t = \frac{\sum_{i: c_i \ge 2} m_{t, i}}{\sum_{i: c_i = 1} m_{t, i}}.
\end{equation}
Here, the numerator represents the count of shared parameters, and the denominator represents the count of unique parameters. Since the sum of these two components constitutes the total updated parameter volume, a higher $\rho_t$ indicates that model learns task $t$ largely within the shared subspace.

\subsection{Rescaling Unique Regions}
\label{sec:rescaling}
Reinforced task vectors with high Overlap-Unique Ratios ($\rho_t$) are likely to suffer greater degradation of task capabilities when their substantial shared regions are averaged with other vectors. Therefore, we proportionally rescale the unique regions of such parts of task vectors to compensate for the performance loss incurred in the shared regions. To achieve this, we calculate a task-specific scaling factor $\lambda_t$ derived from a functional equivalence hypothesis as follows.

Let $\Delta f_t$ denote the functional gain, or task performance gain, induced by the task vector $\boldsymbol{\tau}_t$. We decompose the total gain into shared and unique components based on the overlap statistics defined in Section~\ref{sec:method:probing}:
\begin{equation}
    \Delta f_t = \mathcal{C}_{\text{shared}} + \mathcal{C}_{\text{unique}}, 
\end{equation}
where the gains are modeled as the task vector element $\tau_{t,i}$ weighted by local sensitivity $g_i$, which indicates the contribution coefficient mapping the task
vector element to performance gain. The performance gains are therefore represented as:
$\mathcal{C}_{\text{shared}} = \sum_{i: c_i \ge 2} g_i \tau_{t,i} m_{t,i}, 
    \mathcal{C}_{\text{unique}} = \sum_{i: c_i = 1} g_i \tau_{t,i} m_{t,i}$.
In the merged task vectors, elements in the shared regions are averaged across vectors, and the corresponding task performances are degraded. We model this as a contraction of the effective signal by a coefficient $1-r \in (0, 1)$. To counteract this, we rescale the magnitudes of parameters in unique regions and have a new functional gain expression:
\begin{equation}
    \Delta \hat{f}_t = (1-r) \mathcal{C}_{\text{shared}} + \lambda_t \mathcal{C}_{\text{unique}}.
\end{equation}
We hypothesize that this rescaling operation achieves the \textit{functional equivalence} to have the same performance gain for the task vector $\tau_{t}$ on task $t$: $\Delta \hat{f}_t \approx \Delta f_t$. Under the simplifying assumption that parameter importance is isotropic ($g_i \tau_{t,i} \approx \text{const}$ on average), the ratio of functional contributions $\mathcal{C}_{\text{shared}}/ \mathcal{C}_{\text{unique}}$ reduces to the ratio of parameter counts defined in Eq.~\ref{eq:overlap-unique}. Then by solving \textit{functional equivalence} as our objective, the required amplification satisfies:
\begin{equation}
    \lambda_t - 1 = r \frac{\mathcal{C}_{\text{shared}}}{\mathcal{C}_{\text{unique}}} \approx r \frac{\sum_{i: c_i \ge 2} m_{t,i}}{\sum_{i: c_i = 1} m_{t,i}} = r \rho_t,
\end{equation}
which is approximately proportional to the Overlap-Unique Ratio $\rho_t$. This suggests that tasks with higher overlap require stronger compensation in their unique parameter subspace to counteract the degradation induced by averaging.
However, directly instantiating this proportional relationship may lead to numerical instability when $\rho_t$ is large. To balance signal compensation with stability, we employ a clipped linear scaling rule:
\begin{equation}
    \lambda_t = 1 + r \cdot \mathrm{clip}\left( \rho_t, 0, \alpha \right),
\end{equation}
where $r$ controls the overall amplification strength and $\alpha$ serves as a stable bound. This design preserves the monotonic growth implied by the hypothesis while preventing excessive amplification in high-overlap scenarios.

\subsection{Selective Merging}
\label{sec:method:merging}
Finally, we construct the merged task vector $\boldsymbol{\tau}_{\text{merged}}$ element-wise. Unlike existing merging methods, which effectively divide unique parameters by $N$ (causing signal dilution), our strategy differentiates between shared and unique regions. 
For each parameter index $i$, let $\mathcal{T}_i = \{t \mid m_{t, i} = 1\}$ denote the set of indices of active tasks for that parameter. Note that the cardinality $|\mathcal{T}_i|$ corresponds to the overlap count $c_i$ defined in Section~\ref{sec:method:probing}. The merged element $\tau_{\text{merged}, i}$ is computed as:
\begin{equation}
    \tau_{\text{merged}, i} = 
    \begin{cases} 
        0 & \text{if } |\mathcal{T}_i| = 0, \\
        \lambda_t \cdot \tau_{t, i} & \text{if } \mathcal{T}_i = \{t\}, \\
        \frac{1}{|\mathcal{T}_i|} \sum_{t \in \mathcal{T}_i} \tau_{t, i} & \text{if } |\mathcal{T}_i| \ge 2.
    \end{cases}
    \label{eq:merge_strategy}
\end{equation}
This selective strategy ensures that:
1) \textit{Shared Knowledge} ($|\mathcal{T}_i| \ge 2$) is averaged to balance multi-task capabilities.
2) \textit{Task-Specific Knowledge} ($|\mathcal{T}_i| = 1$) is completely preserved and amplified by $\lambda_t$ to compensate for the contraction of the effective signal in shared regions, explicitly targeting functional equivalence. 3) \textit{No Knowledge} ($|\mathcal{T}_i| = 0$) is set to zero to filter out insignificant parameter fluctuations and ensure that the base model's general capabilities remain undisturbed.

\begin{table*}[]

\resizebox{\textwidth}{!}{

\begin{tabular}{l !{\color{gray!50}\vrule} cccc !{\color{gray!50}\vrule} cccc !{\color{gray!50}\vrule} cccc !{\color{gray!50}\vrule} c}
\toprule

 & \multicolumn{4}{c!{\color{gray!50}\vrule}}{\textbf{Coding}} 
 & \multicolumn{4}{c!{\color{gray!50}\vrule}}{\textbf{Tool Using}} 
 & \multicolumn{4}{c!{\color{gray!50}\vrule}}{\textbf{Memory}} 
 &  \\

 & \multicolumn{2}{c}{LiveBench} & \multicolumn{2}{c!{\color{gray!50}\vrule}}{LiveCodeBench} 
 & \multicolumn{2}{c}{Live} & \multicolumn{2}{c!{\color{gray!50}\vrule}}{Non-Live} 
 & \multicolumn{2}{c}{HotpotQA} & \multicolumn{2}{c!{\color{gray!50}\vrule}}{SQuAD} 
 &  \\

\multirow{-3}{*}{\textbf{Model}} & ACC & UT & ACC & UT & Para & P\_Mul & Para & P\_Mul & 7K & 14K & 32K & 64K & \multirow{-3}{*}{\textbf{Avg}} \\ 
\midrule

\multicolumn{14}{c}{\cellcolor[HTML]{EFEFEF}\textit{Base and Task Models}} \\
Base & 28.35 & 40.87 & 23.43 & 36.42 & \cellcolor[HTML]{FFCCC9}\textbf{56.25} & 41.67 & 68.00 & 55.00 & 60.94 & 50.00 & 64.84 & 58.59 & 48.70 \\
CURE (Coding) & 37.70 & 49.27 & 30.23 & 45.76 & \cellcolor[HTML]{FFCCC9}\textbf{56.25} & 37.50 & 64.00 & 51.50 & 58.59 & 56.25 & 60.94 & 44.22 & 49.35 \\
ToolRL (Tool) & 31.84 & 41.36 & 26.76 & 42.05 & \cellcolor[HTML]{FFCCC9}\textbf{56.25} & 58.33 & 91.00 & 89.00 & 58.59 & 48.44 & 59.38 & 46.95 & 54.16 \\
MemAgent (Memory) & 39.25 & 50.12 & 28.92 & 44.80 & 37.50 & 50.00 & 78.50 & 48.50 & 78.91 & 78.12 & \cellcolor[HTML]{FFCCC9}81.25 & 77.34 & 57.77 \\ 
\midrule

\multicolumn{14}{c}{\cellcolor[HTML]{EFEFEF}\textit{Merged Models}} \\
TA & 38.09 & {\ul 51.62} & {\ul 31.95} & 46.69 & 43.75 & 45.83 & 87.50 & 69.50 & 69.53 & 68.75 & 73.44 & 72.66 & 58.28 \\
Fisher & 36.72 & 48.73 & 30.87 & 45.89 & 43.75 & 41.67 & 86.5 & 63.5 & 60.06 & 49.22 & 58.59 & 60.94 & 52.20 \\
TIES & {\ul 39.25} & 49.88 & 30.63 & 46.32 & 43.75 & 54.17 & 84.00 & 67.50 & 71.88 & \cellcolor[HTML]{FFCCC9}\textbf{82.03} & 75.00 & 75.78 & 60.02 \\
DARE+TA & 37.50 & 48.60 & {\ul 31.95} & 46.69 & {\ul 50.00} & {\ul 62.50} & \cellcolor[HTML]{FFCCC9}\textbf{92.50} & 89.50 & {\ul 76.56} & 76.56 & {\ul 77.34} & 70.31 & 63.33 \\
DARE+TIES & 35.93 & 45.66 & 29.26 & 39.53 & \cellcolor[HTML]{FFCCC9}\textbf{56.25} & 58.33 & {\ul 91.50} & 90.00 & 75.00 & 77.34 & 76.56 & 74.22 & 62.47 \\ 
\midrule

\textbf{RAM} & 38.28 & 49.71 & \cellcolor[HTML]{FFCCC9}\textbf{31.96} & \cellcolor[HTML]{FFCCC9}\textbf{47.72} & \cellcolor[HTML]{FFCCC9}\textbf{56.25} & {\ul 66.67} & 91.00 & \cellcolor[HTML]{FFCCC9}\textbf{91.50} & 75.78 & 75.00 & 74.22 & {\ul 79.69} & {\ul 64.82} \\
\textbf{RAM+} & \cellcolor[HTML]{FFCCC9}\textbf{40.23} & \cellcolor[HTML]{FFCCC9}\textbf{52.57} & 31.60 & {\ul 46.84} & \cellcolor[HTML]{FFCCC9}\textbf{56.25} & \cellcolor[HTML]{FFCCC9}\textbf{70.83} & 90.50 & {\ul 91.00} & \cellcolor[HTML]{FFCCC9}\textbf{79.69} & {\ul 78.13} & \textbf{78.91} & \cellcolor[HTML]{FFCCC9}\textbf{82.03} & \cellcolor[HTML]{FFCCC9}\textbf{66.55} \\ 
\bottomrule
\end{tabular}
}
\caption{
\textbf{Main results of agent merging.} 
We evaluate the capabilities across three domains: Coding (LiveBench, LiveCodeBench), Tool Use (Live, Non-Live), and Memory (RULER-HotpotQA, RULER-SQuAD). 
\textbf{Bold} and \underline{underlined} values denote the best and second-best performance among \textit{merged models}, respectively. 
Cells highlighted in \colorbox[HTML]{FFCCC9}{red} indicate the best performance across \textit{all evaluated models}, including the specialized Task Models. 
}
\label{tab:main_results}
\end{table*}

\begin{figure*}[t]
    \centering
    \begin{subfigure}{1.0\linewidth}
        \centering
        \includegraphics[width=\linewidth]{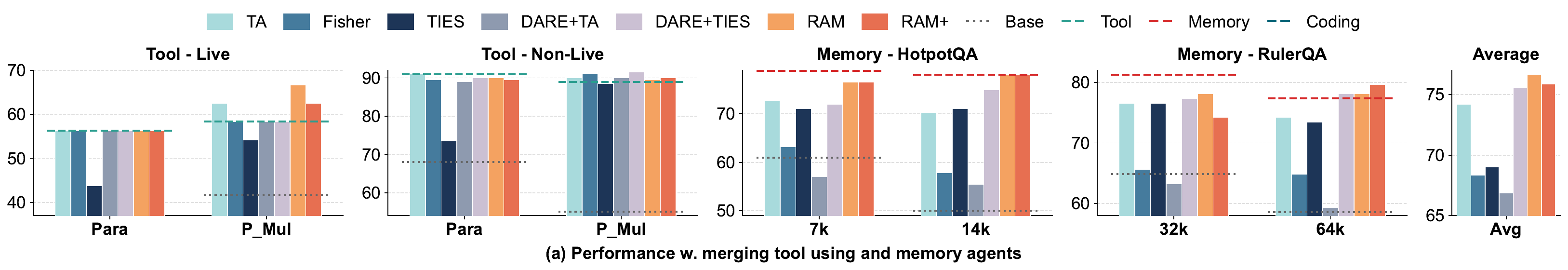}
        \label{fig:sub1}
    \end{subfigure}
    
    \vspace{-0.5cm} 
    
    \begin{subfigure}{1.0\linewidth}
        \centering
        \includegraphics[width=\linewidth]{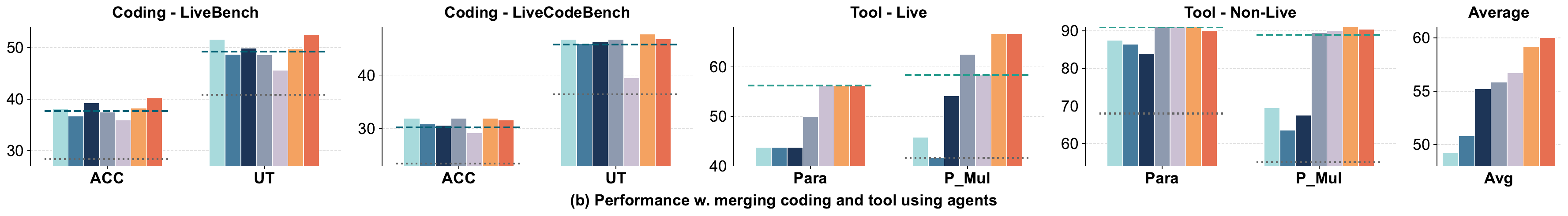}
        \label{fig:sub2}
    \end{subfigure}
    
    \vspace{-0.5cm}
    
    \begin{subfigure}{1.0\linewidth}
        \centering
        \includegraphics[width=\linewidth]{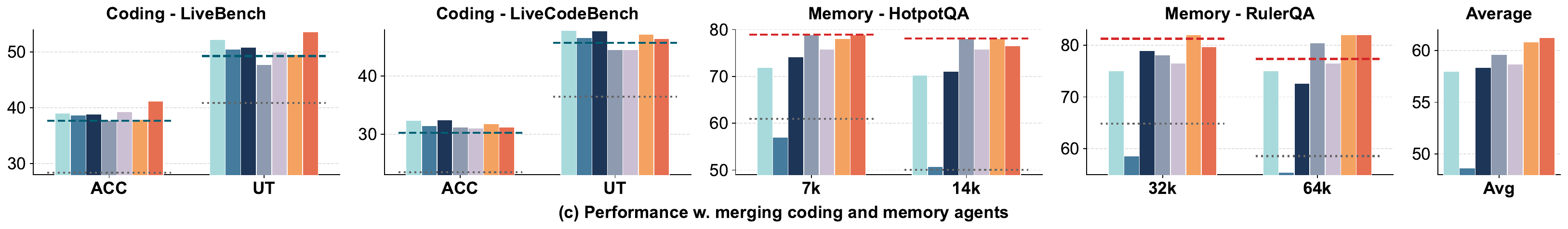}
        \label{fig:sub3}
    \end{subfigure}
    
    \vspace{-0.7cm} 
    \caption{The performances of merging two agents across domains.}
    \label{fig:2model_merge}
\end{figure*}

\section{Experiments}

\subsection{Experimental Setup}
\label{sec:exp:setup}
\paragraph{Baselines} We categorize our baselines into two groups: original specialized agent models and established model merging techniques. \textbf{(i) Original Agent Models:} We utilize \textbf{Qwen2.5-7B-Instruct}~\citep{qwen2} as the shared base model. The task-specific reinforced agents include: \textbf{CURE}~\citep{wang2025cure} for coding, \textbf{ToolRL}~\citep{qian2025toolrl} for tool-use, and \textbf{MemAgent}~\cite{yu2025memagent} for long-context memory. \textbf{(ii) Merging Methods:} We compare our approach against prominent merging strategies: \textbf{Task Arithmetic}~\citep{ilharcoediting}, which linearly combines task vectors; \textbf{Fisher Merging}~\citep{matena2022merging}, which weighs parameters based on Fisher information; \textbf{TIES-Merging}~\citep{yadav2023resolving}, which mitigates parameter interference through trimming and sign consensus; and \textbf{DARE}~\citep{yu2024language}, which randomly drops and rescales the parameters. Following standard practice, we combine DARE with Task Arithmetic and TIES for evaluation. We introduce more details about baselines in Appendix~\ref{app:baseline}.

\paragraph{Evaluations} We evaluate the models across three critical agentic domains: coding, tool-use, and long-context memory. For \textbf{Coding}, we measure generated code pass accuracy (ACC) and unit test pass accuracy (UT) on the LiveBench~\citep{white2025livebench} and LiveCodeBench~\citep{jain2024livecodebench} benchmarks. For \textbf{Tool Use}, we utilize the Berkeley Function Call Leaderboard (BFCL)~\cite{patil2025the}, specifically reporting results on the Live/Non-Live Parallel (Para) and Parallel Multiple (P\_Mul) subsets. For \textbf{Long-Context Memory}, following~\cite{yu2025memagent}, we employ the RULER benchmark~\cite{hsieh2024ruler} to assess performance on long-context tasks, including RULER~\cite{hsieh2024ruler} HotpotQA and SQuAD with 7K, 14K, 32K, and 64K lengths. Additional datasets and detailed evaluation criteria are provided in Appendix~\ref{app:eval}.

\paragraph{Implementations}
For hyperparameters, we set $r=0.1$ and $\alpha=2.0$. We denote our method without task-specific rescaling as \textbf{RAM} and with rescaling as \textbf{RAM+}. RAM is the special case of RAM+ when $r=0$.
We provide the details of pseudocode and agents for merging in Appendix~\ref{app:implementation}.

\subsection{RAM is a Better Fit for Reinforced Models}
\label{sec:exp:main}
Table~\ref{tab:main_results} demonstrates that both RAM and RAM+ consistently outperform all baselines, establishing a new SOTA. 
Specifically, RAM achieves an average score of 64.82 across all tasks, surpassing the strongest baseline DARE (63.33). 
Building on this foundation, RAM+ further pushes the boundary to 66.55 after rescaling unique regions, unlocking synergistic potential among agents where the merged generalist exceeds the capabilities of specialized task agents on most of the evaluations. 
For instance, in the Coding domain, RAM+ surpasses the specialist Coding agent on LiveBench and LiveCodeBench, suggesting that reasoning signals from other tasks enhance coding precision. 
This superiority extends to Tool Use, where RAM+ significantly outperforms the Tool agent in complex parallel scenarios (Live P\_Mul: 70.83 vs. 58.33), and to Long-Context Memory, where it achieves global optimal performance on SQuAD 64k (82.03), beating the dedicated Memory agent (77.34).

\begin{table*}[t]
\resizebox{\textwidth}{!}{
\begin{tabular}{lccccccccccccc}
\toprule
\multirow{3}{*}{\textbf{r}} & \multicolumn{4}{c}{\textbf{Code}}                                   & \multicolumn{4}{c}{\textbf{Tool}}                                 & \multicolumn{4}{c}{\textbf{Memory}}                               & \multirow{3}{*}{\textbf{Avg}} \\
                            & \multicolumn{2}{c}{LiveBench}   & \multicolumn{2}{c}{LiveCodeBench} & \multicolumn{2}{c}{Live}        & \multicolumn{2}{c}{Non-Live}    & \multicolumn{2}{c}{HotpotQA}    & \multicolumn{2}{c}{RulerQA}     &                               \\
                            & ACC            & UT             & ACC             & UT              & Para           & P\_Mul         & Para           & P\_Mul         & 7k             & 14k            & 32k            & 64k            &                               \\ \midrule
\textbf{0.00}               & 38.28          & 49.71          & {\ul 31.96}     & \textbf{47.72}  & {\ul 56.25}    & {\ul 66.67}    & {\ul 91.00}    & \textbf{91.50} & 75.78          & 75.00          & 74.22          & {\ul 79.69}    & 64.82                         \\
\textbf{0.05}               & 37.70          & {\ul 50.49}    & 30.63           & 45.59           & \textbf{62.50} & 62.50          & \textbf{92.00} & \textbf{91.50} & 75.78          & \textbf{79.69} & 75.00          & \textbf{82.03} & 65.45                         \\
\textbf{0.10}               & \textbf{40.23} & \textbf{52.57} & 31.60           & 46.84           & {\ul 56.25}    & \textbf{70.83} & 90.50          & {\ul 91.00}    & \textbf{79.69} & {\ul 78.13}    & 78.91          & \textbf{82.03} & \textbf{66.55}                \\
\textbf{0.15}               & 38.67          & 49.85          & 31.41           & 46.53           & \textbf{62.50} & {\ul 66.67}    & 89.50          & \textbf{91.50} & {\ul 78.12}    & \textbf{79.69} & {\ul 79.69}    & 75.78          & {\ul 65.83}                   \\
\textbf{0.20}               & {\ul 39.45}    & 50.36          & \textbf{32.58}  & {\ul 47.54}     & {\ul 56.25}    & 62.50          & {\ul 91.00}    & 90.50          & {\ul 78.12}    & 78.12          & \textbf{80.47} & 77.34          & 65.35                         \\ \bottomrule
\end{tabular}
}
\caption{
\textbf{Ablation Study.} 
\textbf{Bold} and \underline{underlined} values denote the best and second-best performance.
}
\label{tab:ablation}
\end{table*}

\subsection{Extending Model Combinations}

To evaluate the effectiveness of RAM beyond tri-agent merging, we extend pairwise agent merging experiments with more model combinations across Tool+Memory, Coding+Tool, and Coding+Memory scenarios, as illustrated in Figure~\ref{fig:2model_merge} (details are provided in Appendix~\ref{app:addexp}). Across all three combinations, RAM/RAM+ consistently achieves the highest average performance, demonstrating superior robustness on various combinations compared to baselines. Specifically, in the Coding+Tool setting, RAM+ attains an average score of 60.04, significantly outperforming the strongest baseline DARE+TIES (56.74) and effectively bridging the capability gap that traditional methods like Task Arithmetic and TIES fail to address due to signal dilution. Similarly, in the Tool+Memory and Coding+Memory scenarios, RAM+ maintains dominant performance with average scores of 75.86 and 61.21 respectively, confirming that RAM can be successful in multiple agent combinations.

\subsection{Ablation Study}
\label{sec:ablation}

We ablate our method to RAM ($r=0$) and RAM+, and investigate the sensitivity of our proposed method to the scaling factor $r$. Table~\ref{tab:ablation} presents the ablation results with $r$ varying from $0.00$ to $0.20$. Note that when $r=0$, the method represents RAM. As observed, the model's overall performance (Avg) exhibits a trend of initially increasing and then decreasing. The performance peaks at $r=0.10$, achieving the highest average score of 66.55. Specifically, setting $r=0.10$ (RAM+) yields the best or second-best results across the majority of metrics, particularly showing significant gains in LiveBench (Coding) and HotpotQA (Memory) compared to RAM ($r=0.00$). However, further increasing $r$ beyond $0.10$ leads to diminishing returns, with the average score dropping to $65.35$ at $r=0.20$. This suggests that while a moderate scaling factor effectively enhances task-specific capabilities, an excessively large $r$ may disrupt the general knowledge of the merged model. 

\subsection{Extending Architecture and Domains}

\begin{figure}[t]  
    \centering
    \includegraphics[width=\linewidth]{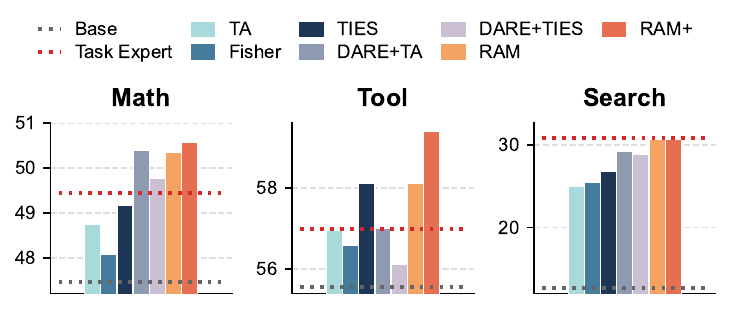} 
    
    \caption{Merging results for RL agents trained from Llama3.2-3B-Instruction base model.}
    \label{fig:llama}
\end{figure}

Besides the agents trained from Qwen architecture, we extend the experiment to Llama architecture and additional domains. We choose Llama3.2-3B~\cite{grattafiori2024llama} as the base model, and choose models trained from it via RL: search agent ZeroSearch~\cite{sun2025zerosearch}, math reasoning agent~\cite{zhao2025learning}, and tool-using agent ToolRL~\cite{qian2025toolrl}. The evaluation details are provided in Appendix~\ref{app:llama_setting}.
Figure~\ref{fig:llama} illustrates that, consistent with our findings on the Qwen, RAM and RAM+ demonstrate superior performance across multiple agentic domains, consistently outperforming baselines. Notably, they achieve positive synergy in both Math and Tool domains, where the merged generalist surpasses the performance of the original specialized agents. For instance, in the Tool domain, RAM+ exhibits a significant margin over the specialist, suggesting that reasoning capabilities from Math and Search agents synergize to enhance tool-use. In the Search domain, RAM/RAM+ successfully retain original capability, whereas baselines show notable regression. These results confirm that the heterogeneity of reinforced task vectors is a general property, and RAM effectively addresses this by preserving task-specific specialized knowledge independent of model scale and architecture.

\paragraph{Additional Experiments} Besides the above experiments, we further provide the instruction following evaluation to assess the forgetting after merging in Appendix~\ref{app:ifeval}; Merging efficiency comparison in Appendix~\ref{app:merging_efficiency}; Evaluation on the additional tasks in Appendix~\ref{app:add_tasks}; Additional rescaling strategy performance in Appendix~\ref{app:scaling_derivation}.

\section{Conclusion}

In this work, we address the critical challenge of merging agents fine-tuned via RL, identifying a fundamental mismatch between standard merging techniques designed for dense SFT updates and the sparse, heterogeneous nature of on-policy RL task vectors. We demonstrate that treating global task vectors equally in previous methods in this setting leads to signal dilution of task-specific capabilities. To bridge this gap, we propose Reinforced Agent Merging (RAM), a method that explicitly disentangles shared and unique parameter update regions and applies a distribution-based rescaling strategy to preserve specialized behaviors. Extensive evaluations across multiple agentic domains and model architectures show that RAM significantly outperforms existing baselines, achieving SOTA results and surpassing the original specialists as a unified generalist on most tasks. 

\clearpage
\section*{Limitations}
\label{sec:limitations}

While Reinforced Agent Merging (RAM) effectively mitigates signal dilution for RL-trained agents, our current study has the following limitations. First, our experiments focus on merging a common number of agents; as the number of agents scales significantly, the probability of parameter collision in the shared subspace increases, potentially requiring more complex conflict resolution strategies beyond simple averaging. Second, the derivation of our rescaling factor relies on an isotropic assumption of parameter importance, which, while empirically robust, does not explicitly account for element-wise curvature information that could offer finer-grained control at a higher computational cost. Third, although we identified a default hyperparameter configuration that generalizes well across Qwen and Llama architectures, optimal performance on agents trained with fundamentally different data or modalities may require task-specific tuning. Finally, our evaluation is primarily conducted on 3B and 7B parameter models; verifying whether the sparsity hypothesis and RAM's efficacy persist in massive-scale models (70B+) remains an open question for future research.


\bibliography{ref}

\clearpage
\appendix

\section{Implementation Details}
\label{app:implementation}

\subsection{Pseudocode}
\label{app:Pseudocode}

Here we provide the Pseudocode of RAM in Algorithm~\ref{alg:ram}.

\begin{algorithm}[]
\caption{Reinforced Agent Merging (RAM+)}
\label{alg:ram}
\footnotesize 
\begin{algorithmic}[1]
\raggedright 

\REQUIRE Task vectors $\{\boldsymbol{\tau}_t\}_{t=1}^N$; threshold $\epsilon$; rescale strength $r$; clip bound $\alpha$
\ENSURE Merged task vector $\boldsymbol{\tau}_{\text{merged}}$

\STATE \algComment{Stage 1: Probing Vector Distribution (Sec.~\ref{sec:method:probing})}
\FOR{$t=1$ \TO $N$}
    \STATE $\mathbf{m}_t \leftarrow \mathbb{I}(|\boldsymbol{\tau}_t| > \epsilon)$ 
\ENDFOR
\STATE $\mathbf{c} \leftarrow \sum_{t=1}^N \mathbf{m}_t$ 

\STATE \algComment{Stage 2: Rescaling Unique Regions (Sec.~\ref{sec:rescaling})}
\FOR{$t=1$ \TO $N$}
    \STATE $N_{\text{shared}} \leftarrow \sum_{i: c_i \ge 2} m_{t,i}$
    \STATE $N_{\text{unique}} \leftarrow \sum_{i: c_i = 1} m_{t,i}$
    \STATE $\rho_t \leftarrow N_{\text{shared}} / N_{\text{unique}}$ 
    \STATE $\lambda_t \leftarrow 1 + r \cdot \text{clip}(\rho_t, 0, \alpha)$ 
\ENDFOR

\STATE \algComment{Stage 3: Selective Merging (Sec.~\ref{sec:method:merging})}
\FOR{each parameter index $i$}
    \STATE $\mathcal{T}_i \leftarrow \{t \mid m_{t,i} = 1\}$

    \IF{$|\mathcal{T}_i| = 0$}
        \STATE $\tau_{\text{merged}, i} \leftarrow 0$
    \ELSIF{$|\mathcal{T}_i| = 1$}
        \STATE Let $t$ be the unique element in $\mathcal{T}_i$
        \STATE $\tau_{\text{merged}, i} \leftarrow \lambda_t \cdot \tau_{t, i}$ 
    \ELSE
        \STATE $\tau_{\text{merged}, i} \leftarrow \frac{1}{|\mathcal{T}_i|} \sum_{t \in \mathcal{T}_i} \tau_{t, i}$ 
    \ENDIF
\ENDFOR

\RETURN $\boldsymbol{\tau}_{\text{merged}}$
\end{algorithmic}
\end{algorithm}

\subsection{Details of Reinforced Task Agents}
\label{app:agent_details}

In this section, we provide the detailed specifications and sources for the reinforced agentic models and base models used in our experiments. All models are publicly available on Hugging Face.

\paragraph{Qwen2.5-7B-Instruction Series:} We utilize the following agents initialized from the Qwen2.5-7B-Instruct:

\vspace{0.5em}

\noindent \textit{Base Model} (Qwen2.5-7B-Instruct)~\cite{qwen2} \\
\href{https://huggingface.co/Qwen/Qwen2.5-7B-Instruct}{\footnotesize\texttt{\hfemoji\ Qwen2.5-7B-Instruct}}

\vspace{0.4em}

\noindent \textit{Coding Agent} (CURE)~\cite{wang2025cure} \\
\href{https://huggingface.co/Gen-Verse/ReasonFlux-Coder-7B}{\footnotesize\texttt{\hfemoji\ ReasonFlux-Coder-7B}}

\vspace{0.4em}

\noindent \textit{Tool Agent} (ToolRL)~\cite{qian2025toolrl} \\
\href{https://huggingface.co/emrecanacikgoz/Qwen2.5-7B-Instruct-ToolRL-grpo-cold}{\footnotesize\texttt{\hfemoji\ Qwen2.5-7B-Instruct-ToolRL-grpo-cold}}

\vspace{0.4em}

\noindent \textit{Memory Agent} (MemAgent)~\cite{yu2025memagent} \\
\href{https://huggingface.co/BytedTsinghua-SIA/RL-MemoryAgent-7B}{\footnotesize\texttt{\hfemoji\ RL-MemoryAgent-7B}}

\vspace{0.4em}

\noindent \textit{Search Agent} (ZeroSearch)~\cite{sun2025zerosearch} \\
\href{https://huggingface.co/Alibaba-NLP/ZeroSearch_google_V2_Qwen2.5_7B_Instruct}{\footnotesize\texttt{\hfemoji\ ZeroSearch\_google\_V2\_Qwen2.5\_7B\_Instruct}}

\vspace{0.4em}

\noindent \textit{Tool Integrated Reasoning Agent} (AutoTIR)~\cite{wei2025autotir} \\
\href{https://huggingface.co/Weiyifan/AutoTIR-Qwen2.5-7B-Instruct}{\footnotesize\texttt{\hfemoji\ AutoTIR-Qwen2.5-7B-Instruct}}

\paragraph{Llama-3.2-3B-Instruction Series} To verify generalization across architectures, we utilize the following agents based on Llama-3.2-3B-Instruct:

\vspace{0.5em}

\noindent \textit{Base Model} (Llama-3.2-3B-Instruct)~\cite{grattafiori2024llama} \\
\href{https://huggingface.co/unsloth/Llama-3.2-3B-Instruct}{\footnotesize\texttt{\hfemoji\ Llama-3.2-3B-Instruct}}

\vspace{0.4em}

\noindent \textit{Math Agent} (GRPO-Math)~\cite{guo2025deepseek} \\
\href{https://huggingface.co/sunblaze-ucb/Llama-3.2-3B-Instruct-GRPO-MATH-1EPOCH}{\footnotesize\texttt{\hfemoji\ Llama-3.2-3B-Instruct-GRPO-MATH-1EPOCH}}

\vspace{0.4em}

\noindent \textit{Tool Agent} (ToolRL)~\cite{qian2025toolrl} \\
\href{https://huggingface.co/chengq9/ToolRL-Llama3.2-3B}{\footnotesize\texttt{\hfemoji\ ToolRL-Llama3.2-3B}}

\vspace{0.4em}

\noindent \textit{Search Agent} (ZeroSearch)~\cite{sun2025zerosearch} \\
\href{https://huggingface.co/Alibaba-NLP/ZeroSearch_google_V2_Llama_3.2_3B_Instruct}{\footnotesize\texttt{\hfemoji\ ZeroSearch\_google\_V2\_Llama\_3.2\_3B\_Instruct}}

\section{Evaluation Details}
\label{app:eval}

\subsection{Coding Evaluation}
\label{subsec:coding_eval}

Following the evaluation setting established in \citet{wang2025cure}, we conduct a comprehensive evaluation of coding capabilities across five widely adopted coding benchmarks.

\paragraph{Datasets.}
We utilize \textbf{LiveBench}~\cite{white2025livebench} (standard test set), \textbf{MBPP}~\citep{austin2021program} (standard test set), and \textbf{LiveCodeBench}~\citep{jain2024livecodebench} (Version 2, 511 problems). For competition-level tasks, we include \textbf{CodeContests}~\citep{li2022competition}, filtering for tasks with difficulty $\le 2$ and utilizing a held-out split of 200 examples. Additionally, we use \textbf{CodeForces}~\citep{penedo2025codeforces}, comprising 500 randomly sampled examples distinct from CodeContests.

\paragraph{Evaluation Protocol.}
To ensure consistency, all datasets are standardized to the stdio format. Functional inputs from LiveBench, LiveCodeBench, and MBPP are converted by placing variables on separate lines and flattening lists. For verification, we use official ground-truth solutions for CodeContests and MBPP. For the remaining datasets (CodeForces, LiveCodeBench, LiveBench), we utilize high-quality reference solutions generated by QwQ-32B (via Best-of-3 sampling).
We employ vLLM~\citep{kwon2023efficient} for generation. Following standard practices, sampling parameters are set to temperature $T=1.0$, top-$p=0.95$. We report performance using pass accuracy, including code pass (ACC) and unit test pass (UT) (Pass@1) and Best-of-N (BoN, N=4) metrics. 

\subsection{Long-Context Memory Evaluation}
\label{subsec:memory_eval}

To rigorously assess the long-context memory capabilities of our agents, we adopt the evaluation protocol from the RULER benchmark~\citep{hsieh2024ruler}, strictly following the data synthesis configurations established in \citet{yu2025memagent}. We specifically select RULER-HotpotQA and RULER-SQuAD as our primary benchmarks to evaluate multi-hop reasoning and precise fact retrieval.

\paragraph{Datasets.}
We utilize the following two tasks adapted for long-context memory evaluation:
\begin{itemize}
    \item \textbf{RULER-HotpotQA:} This task serves as a robust testbed for \textit{multi-hop reasoning}. In this setup, multiple "golden paragraphs" containing necessary evidence are embedded within a vast amount of distractor content (the haystack). The model must effectively identify and synthesize these scattered pieces of evidence from its memory to correctly answer complex questions.
    \item \textbf{RULER-SQuAD:} Adapted from the SQuAD dataset, this task evaluates precise reading comprehension. Ground-truth passages are inserted into long distractor texts, requiring the model to maintain high fidelity to specific facts over extended sequences. This tests the agent's ability to accurately recall specific instructions or details without hallucination.
\end{itemize}

\paragraph{Evaluation Protocol.}
Consistent with \citet{yu2025memagent}, we synthesize test samples with varying context lengths to stress-test memory capacity with different context lengths (8K-128K for SQuAD and 7K-896K for HotPotQA). The primary evaluation metric is the Substring Exact Match (sub\_em) of the generated answers. High accuracy in this setting demonstrates that the merged agent successfully retains critical task-specific memory capabilities and can effectively filter out noise (distractors) inherent in long-context processing.

\subsection{Tool Use Evaluation}
\label{subsec:tool_eval}

To comprehensively evaluate the tool-use (function calling) capabilities of our agents, we employ the Berkeley Function Calling Leaderboard (BFCL)~\citep{patil2025the}, widely recognized as the standard benchmark for assessing LLM agentic behaviors. We specifically use the \textit{Live} and \textit{Non-Live} datasets to measure performance across both real-world and synthetic scenarios.

\paragraph{Datasets.}
We utilize the following two subsets to assess distinct dimensions of function calling:
\begin{itemize}
    \item \textbf{Non-Live Dataset (Synthetic \& Curated):} Derived from BFCL V1, this subset consists of expert-curated synthetic tasks designed to test fundamental logic across various languages (Python, Java, JavaScript) and SQL. It evaluates the model's adherence to precise instructions in controlled environments. The tasks of Non-Live datasets include: \texttt{Multiple, Parallel, Relevance, Simple, Parallel\_multiple and Irrelevance}.
    \item \textbf{Live Dataset (Real-World \& Crowdsourced):} Introduced in BFCL V2, this subset comprises user-contributed examples from real-world agent interactions. Unlike the Non-Live set, these samples are diverse and noisy, involving complex APIs with nested parameters. This benchmark specifically challenges the model's robustness in handling ambiguous queries and detecting function irrelevance, including seven tasks: \texttt{Multiple, Irrelevance, Simple\_java, Simple\_javascript, Parallel\_multiple, Parallel and Simple\_python}.
\end{itemize}

\paragraph{Evaluation Protocol.}
To ensure robust evaluation, we utilize the Abstract Syntax Tree (AST) matching method provided by the BFCL framework. Unlike simple string matching, AST evaluation parses generated function calls into syntax trees to structurally verify argument permutations and formatting variations while enforcing strict type correctness. We report accuracy for both Live and Non-Live splits, with a particular focus on the challenging \textit{Parallel} and \textit{Parallel Multiple} categories to demonstrate advanced planning capabilities.

\subsection{Evaluation Details for Llama-based Agents}
\label{app:llama_setting}
We evaluate agents trained from LLama3.2-3B-Instruction on three domains: math, search, and tool-use. For the math domain, we evaluate the model on GSM8K~\cite{cobbe2021training} and MATH500~\cite{hendrycks2measuring} datasets, provided by LM-Evaluation-Harness~\cite{eval-harness}. For the search domain, we follow the evaluation setting provided in ZeroSearch~\cite{sun2025zerosearch} and evaluate the agent on NQ~\cite{kwiatkowski2019natural} and 2WikiMultiHopQA~\cite{ho2020constructing}. For tool-use, the setting is the same as Section~\ref{sec:exp:setup}. We take the average score across tasks for each domain for evaluation.

\section{Additional Experiments}
\label{app:addexp}

\subsection{Additional Reinforced Task Vector Analysis}
\label{app:add_behavior}
\begin{figure}[t] 
    \centering
    \includegraphics[width=1.0\linewidth]{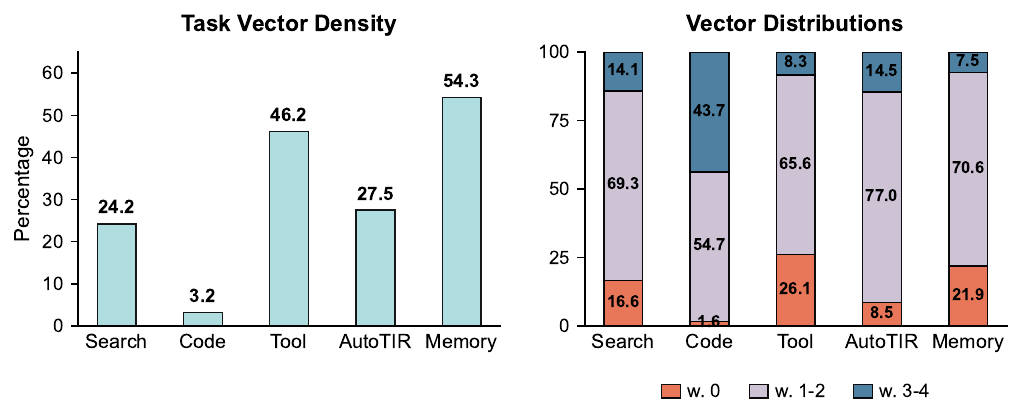}
    \caption{Additional distribution analysis for sparse reinforced task vectors trained from Qwen2.5-7B-Instruction. \textbf{Left}: Density (1-Sparsity) of task vectors varies between agent models. \textbf{Right}: The number of overlaps with other task vectors.}
    \label{fig:motivation2}
    
\end{figure}

\begin{figure}[t]
    \centering
    \includegraphics[width=1.0\linewidth]{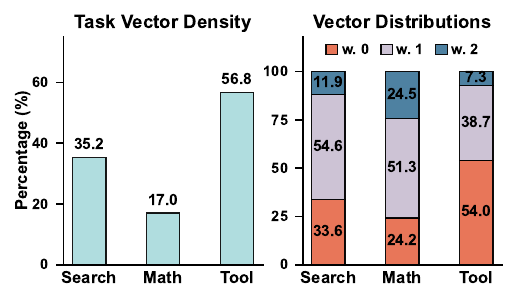}
    \caption{The task vector analysis for agents trained from Llama3.2-3B-Instruction via RL. \textbf{Left}: Density (1-Sparsity) of task vectors varies between tool-using agent, web search agent, and the math reasoning agent models. \textbf{Right}: The number of overlaps with other task vectors.}
    \label{fig:overlap-analysis-llama}
\end{figure}

To further verify heterogeneity in reinforced task vectors introduced by Section~\ref{sec:behavior_analysis}, we extend the number, domains, and architecture of the reinforced agents. 

First, we include extra agents, web search agent ZeroSearch~\cite{sun2025zerosearch} and tool-integrated reasoning agent AutoTIR~\cite{wei2025autotir}, which are both RL-trained from Qwen2.5-Instruct-7B. Figure~\ref{fig:motivation2} shows that when including five agents specialized in multiple domains together, the heterogeneity in sparsity and distribution remains significant. Specifically, the sparsity of task vectors spans a wide spectrum, ranging from merely 3.2\% for the Code agent to 54.3\% for the Memory agent. The overlap analysis further reveals distinct behaviors: the Code agent is highly entangled with others, with 43.7\% of its changed parameters shared among 3 or 4 other agents (w. 3-4). In contrast, agents like Tool and Memory maintain higher independence, with unique parameter ratios (w. 0) of 26.1\% and 21.9\%, respectively.

Second, we extend the experiment to Llama architecture and additional domains. We choose Llama3.2-3B-Instruction~\cite{grattafiori2024llama} as the base model, and choose models trained from it via RL: search agent ZeroSearch~\cite{sun2025zerosearch}, math reasoning agent~\cite{zhao2025learning}, and tool-using agent ToolRL~\cite{qian2025toolrl}. Figure~\ref{fig:overlap-analysis-llama} demonstrates that similar heterogeneity in task vectors persists across different model architectures. As shown in the left panel, the sparsity of task vectors varies significantly, ranging from 17.0\% for the Math agent to 56.8\% for the Tool agent. The overlap distribution (right panel) further highlights this diversity: the Tool agent modifies a large proportion of unique parameters (54.0\%), whereas the Math agent shares the majority of its updates with other tasks, with only 24.2\% of its modified parameters being unique. This confirms that the diverse characteristics of reinforced task vectors are consistent across different base models and task domains.

\subsection{Instruction Following Evaluation} 
\label{app:ifeval}

\begin{table*}[t]
\centering
\resizebox{1.0\textwidth}{!}{%
\begin{tabular}{lcccccccc}
\toprule
& \multicolumn{4}{c}{\textbf{Qwen2.5-7B-Instruct}} & \multicolumn{4}{c}{\textbf{Llama-3.2-3B-Instruction}} \\
\cmidrule(lr){2-5} \cmidrule(lr){6-9}
& \multicolumn{2}{c}{\textbf{Instruction Acc (\%)}} & \multicolumn{2}{c}{\textbf{Prompt Acc (\%)}} & \multicolumn{2}{c}{\textbf{Instruction Acc (\%)}} & \multicolumn{2}{c}{\textbf{Prompt Acc (\%)}} \\
\textbf{Model} & \textbf{Loose} & \textbf{Strict} & \textbf{Loose} & \textbf{Strict} & \textbf{Loose} & \textbf{Strict} & \textbf{Loose} & \textbf{Strict} \\
\midrule
\multicolumn{9}{c}{\cellcolor[HTML]{EFEFEF}\textit{Base and Task Experts}} \\
Base Model & 69.18 & 64.39 & 58.96 & 53.23 & \textbf{69.90} & \textbf{63.31} & \textbf{59.15} & \textbf{51.20} \\
Code/Math & \res{73.14}{3.96} & \res{68.11}{3.72} & \res{62.48}{3.52} & \res{58.75}{5.52} & \res{68.59}{-1.31} & \res{61.99}{-1.32} & \res{57.49}{-1.66} & \res{48.43}{-2.77} \\
Tool & \res{71.82}{2.64} & \res{66.07}{1.68} & \res{61.55}{2.59} & \res{54.34}{1.11} & \res{68.11}{-1.79} & \res{61.87}{-1.44} & \res{57.12}{-2.03} & \res{49.35}{-1.85} \\
Memory/Search & \res{69.78}{0.60} & \res{66.07}{1.68} & \res{58.41}{-0.55} & \res{53.60}{0.37} & \res{67.63}{-2.27} & \res{60.67}{-2.64} & \res{55.45}{-3.70} & \res{46.58}{-4.62} \\
\midrule
\multicolumn{9}{c}{\cellcolor[HTML]{EFEFEF}\textit{Merged Models}} \\
Task Arithmetic (TA) & \res{70.74}{1.56} & \res{65.71}{1.32} & \res{58.90}{-0.06} & \res{53.23}{0.00} & \res{68.82}{-1.08} & \res{61.87}{-1.44} & \res{57.12}{-2.03} & \res{49.72}{-1.48} \\
Fisher & \textbf{\res{72.06}{2.88}} & \textbf{\res{66.91}{2.52}} & \textbf{\res{61.18}{2.22}} & \textbf{\res{55.08}{1.85}} & \res{67.87}{-2.03} & \res{61.75}{-1.56} & \res{55.82}{-3.33} & \res{49.17}{-2.03} \\
TIES & \res{71.34}{2.16} & \res{67.39}{3.00} & \res{59.52}{0.56} & \res{55.64}{2.41} & \res{58.75}{-11.15} & \res{58.87}{-4.44} & \res{45.84}{-13.31} & \res{46.21}{-4.99} \\
DARE+TA & \res{70.38}{1.20} & \res{65.95}{1.56} & \res{58.41}{-0.55} & \res{53.05}{-0.18} & \res{59.11}{-10.79} & \res{54.20}{-9.11} & \res{46.21}{-12.94} & \res{41.04}{-10.16} \\
DARE+TIES & \res{69.42}{0.24} & \res{65.11}{0.72} & \res{57.67}{-1.29} & \res{52.31}{-0.92} & \res{58.63}{-11.27} & \res{53.48}{-9.83} & \res{45.66}{-13.49} & \res{39.56}{-11.64} \\
\midrule
\textbf{RAM (Ours)} & \res{70.62}{1.44} & \res{65.95}{1.56} & \res{59.70}{0.74} & \res{53.23}{0.00} & \res{67.39}{-2.51} & \res{61.39}{-1.92} & \res{55.45}{-3.70} & \res{47.87}{-3.33} \\
\textbf{RAM+ (Ours)} & \res{69.18}{0.00} & \res{64.75}{0.36} & \res{57.86}{-1.10} & \res{51.94}{-1.29} & \res{66.19}{-3.71} & \res{61.27}{-2.04} & \res{53.97}{-5.18} & \res{47.50}{-3.70} \\
\bottomrule
\end{tabular}%
}
\caption{Evaluation of General Capabilities on IFEval benchmark. We report \textbf{Instruction Accuracy} and \textbf{Prompt Accuracy} under both \textbf{Loose} and \textbf{Strict} criteria. Values in parentheses denote the absolute change relative to the Base Model. Green values indicate improvement, while orange values indicate regression. RAM demonstrates superior robustness compared to TIES/DARE, especially on the smaller Llama-3.2-3B-Instruction model.}
\label{tab:ifeval_results}
\end{table*}

A primary concern in model merging, particularly when combining agents fine-tuned via RL on disparate domains, is the potential degradation of the base model's general instruction following capabilities (i.e., catastrophic forgetting). To rigorously evaluate whether RAM compromises the model's ability to follow general instructions while pursuing task specialization, we conducted evaluations on the \textbf{IFEval} (Instruction Following Evaluation) benchmark \cite{zhou2023instruction} provided by LM-Evaluation-Harness~\cite{eval-harness}. We report results across four metrics: Instruction Accuracy and Prompt Accuracy, under both Loose and Strict evaluation criteria.
The results are presented in Table~\ref{tab:ifeval_results}. We evaluated two sets of models: \begin{itemize} \item \textbf{Qwen2.5-7B-Instruct:} The primary setting used in the main paper, trained via RL to obtain the Coding, Tool, and Memory agents. \item \textbf{Llama-3.2-3B-Instruction:} To test the generalization of our method on a different architecture and scale, we use the fine-tuned Math, Tool, and Search agents based on Llama-3.2-3B-Instruction. \end{itemize}

On the Qwen-based agents, RAM not only retains the general capabilities of the Base model but explicitly outperforms it across most metrics (e.g., +1.44 in Loose Instruction Accuracy and +1.56 in Strict Instruction Accuracy). This suggests that the specialized reasoning circuits preserved by RAM's disjoint merging strategy can positively transfer to general instruction following. RAM+ shows a slight trade-off, generally maintaining parity with the base model on instruction-level metrics while incurring minor regressions in prompt-level accuracy.
Merging on smaller Llama-based models is inherently more challenging due to limited parameter redundancy. While all merging methods exhibit some regression compared to the Base model, RAM demonstrates superior stability with less forgetting on instruction following. Notably, baseline methods like TIES and DARE+TIES suffer from severe performance collapse, dropping over 10 percentage points (e.g., -11.15 in Loose Instruction Accuracy). Even in strict evaluation, TIES fails to maintain robustness. In contrast, RAM avoids this collapse, showing significantly smaller regressions (approx. 2-3) and proving it is a much safer merging strategy for smaller architectures compared to aggressive trimming methods.

\subsection{Merging Efficiency}
\label{app:merging_efficiency}
\begin{figure}[t]  
    \centering
    \includegraphics[width=\linewidth]{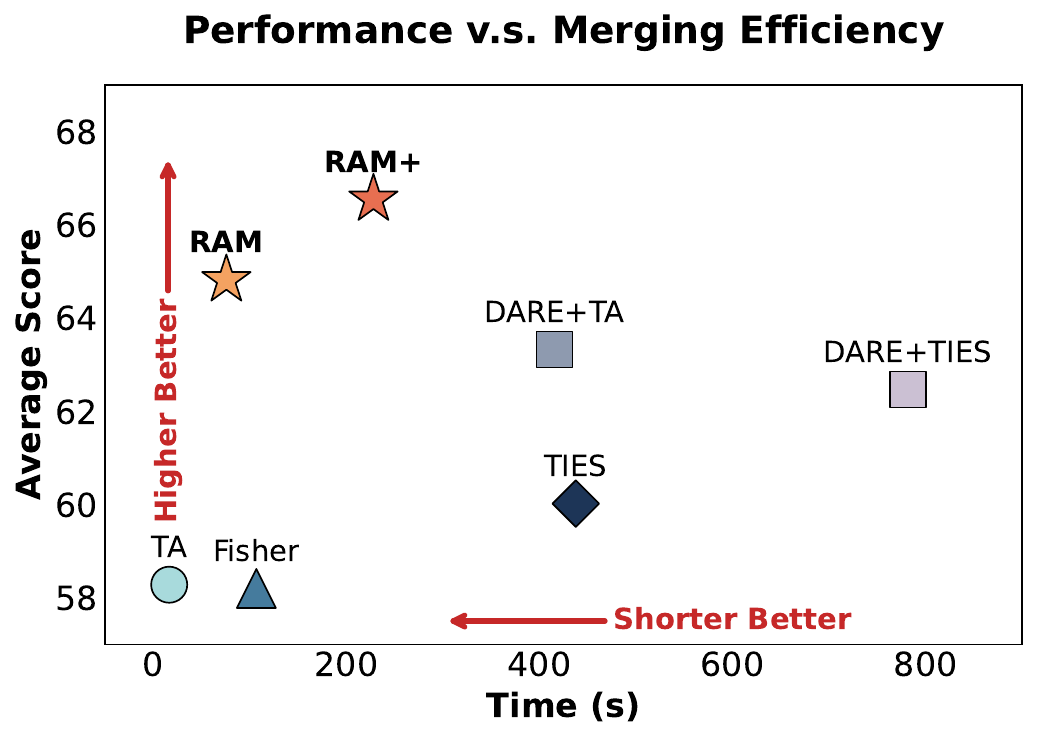} 
    
    \caption{RAM/RAM+ demonstrates a superior trade-off between merging time and average score compared to the baseline method.}
    \label{fig:pareto}
\end{figure}

Beyond merged performance, computational efficiency is another important factor for practical model merging. Figure~\ref{fig:pareto} illustrates the comparison between merging time (in seconds) and the average score across benchmarks. Current baselines exhibit a clear dichotomy: methods like TA and Fisher are computationally efficient (<110s) but suffer from suboptimal performance (around 58.2), while complex methods like TIES and DARE variants achieve better scores at the cost of significant computational overhead (>400s). In contrast, our proposed methods occupy the Pareto frontier of the efficiency-performance landscape. Specifically, RAM achieves a remarkable score of 64.82 in just 75.4 seconds, surpassing DARE+TA in performance while offering a 5.5$\times$ speedup. Even our more intensive variant, RAM+, establishes a new SOTA performance (66.55) while remaining significantly faster than both TIES and DARE. These results demonstrate that RAM effectively identifies critical parameters for merging processing without the extensive computational redundancy found in previous SOTAs.

\subsection{Detailed Numerical Results for Pairwise Agent Merging}
\label{app:exp:2model}

\begin{table*}[]
\centering
\resizebox{0.75\textwidth}{!}{

\begin{tabular}{lccccccccc}
\toprule
                                 & \multicolumn{4}{c}{\textbf{Code}}                                                                                                                                 & \multicolumn{4}{c}{\textbf{Tool}}                                                                                                                                 &                                        \\
                                 & \multicolumn{2}{c}{LiveBench}                                                   & \multicolumn{2}{c}{LiveCodeBench}                                               & \multicolumn{2}{c}{Live}                                                        & \multicolumn{2}{c}{Non-Live}                                                    &                                        \\
\multirow{-3}{*}{\textbf{Model}} & ACC                                    & UT                                     & ACC                                    & UT                                     & Para                                   & P\_Mul                                 & Para                                   & P\_Mul                                 & \multirow{-3}{*}{\textbf{Avg}}         \\ \midrule
\multicolumn{10}{c}{\cellcolor[HTML]{EFEFEF}\textit{Base and Task Models}}                                                                                                                                                                                                                                                                                                                                        \\
Base                             & 28.35                                  & 40.87                                  & 23.43                                  & 36.42                                  & {\ul 56.25}                            & 41.67                                  & 68.00                                  & 55.00                                  & 43.75                                  \\
CURE (Coding)                           & 37.70                                  & 49.27                                  & 30.23                                  & 45.76                                  & {\ul 56.25}                            & 37.50                                  & 64.00                                  & 51.50                                  & 46.53                                  \\
ToolRL (Tool)                             & 31.84                                  & 41.36                                  & 26.76                                  & 42.05                                  & {\ul 56.25}                            & 58.33                                  & \cellcolor[HTML]{FFCCC9}91.00          & 89.00                                  & 54.57                                  \\ \midrule
\multicolumn{10}{c}{\cellcolor[HTML]{EFEFEF}\textit{Merged Models}}                                                                                                                                                                                                                                                                                                                                               \\
TA                               & {\ul 37.11}                            & 49.09                                  & 29.55                                  & 45.83                                  & 50.00                                  & 37.50                                  & 86.00                                  & 59.00                                  & 49.26                                  \\
Fisher                           & 36.72                                  & 50.29                                  & 30.68                                  & \cellcolor[HTML]{FFCCC9}\textbf{48.18} & 50.00                                  & 37.50                                  & 88.50                                  & 64.50                                  & 50.80                                  \\
TIES                             & 35.74                                  & 46.50                                  & {\ul 31.21}                            & 44.48                                  & {\ul 56.25}                            & 54.17                                  & 83.00                                  & {\ul 90.50}                            & 55.23                                  \\
DARE+TA                          & 35.74                                  & 47.40                                  & 30.04                                  & 44.62                                  & {\ul 56.25}                            & {\ul 58.33}                            & 90.00                                  & 84.50                                  & 55.86                                  \\
DARE+TIES                        & 35.54                                  & 45.79                                  & {\ul 31.21}                            & 44.48                                  & {\ul 56.25}                            & \cellcolor[HTML]{FFCCC9}\textbf{66.67} & 88.50                                  & 85.50                                  & 56.74                                  \\ \midrule
\textbf{RAM}                     & \cellcolor[HTML]{FFCCC9}\textbf{39.45} & \cellcolor[HTML]{FFCCC9}\textbf{51.42} & {\ul 31.21}                            & 46.51                                  & {\ul 56.25}                            & \cellcolor[HTML]{FFCCC9}\textbf{66.67} & \cellcolor[HTML]{FFCCC9}\textbf{91.00} & \cellcolor[HTML]{FFCCC9}\textbf{91.00} & {\ul 59.19}                            \\
\textbf{RAM+}                    & \cellcolor[HTML]{FFCCC9}\textbf{39.45} & {\ul 51.10}                            & \cellcolor[HTML]{FFCCC9}\textbf{32.53} & {\ul 47.55}                            & \cellcolor[HTML]{FFCCC9}\textbf{62.50} & \cellcolor[HTML]{FFCCC9}\textbf{66.67} & {\ul 90.50}                            & 90.00                                  & \cellcolor[HTML]{FFCCC9}\textbf{60.04} \\ \bottomrule
\end{tabular}
}
\caption{
\textbf{Detailed results of model merging with code and tool using agents.} 
\textbf{Bold} and \underline{underlined} values denote the best and second-best performance among \textit{merged models}, respectively. 
Cells highlighted in \colorbox[HTML]{FFCCC9}{red} indicate the best performance across \textit{all evaluated models}, including the Task Models.
}
\label{tab:code-tool}
\end{table*}

\begin{table*}[]
\centering
\resizebox{0.75\textwidth}{!}{
\begin{tabular}{lccccccccc}
\toprule
                                 & \multicolumn{4}{c}{\textbf{Tool}}                                                                                                                                 & \multicolumn{4}{c}{\textbf{Memory}}                                                                                                             &                                        \\
                                 & \multicolumn{2}{c}{Live}                                                        & \multicolumn{2}{c}{Non-Live}                                                    & \multicolumn{2}{c}{HotpotQA}                                           & \multicolumn{2}{c}{RulerQA}                                            &                                        \\
\multirow{-3}{*}{\textbf{Model}} & Para                                   & P\_Mul                                 & Para                                   & P\_Mul                                 & 7k                            & 14k                                    & 32k                           & 64k                                    & \multirow{-3}{*}{\textbf{Avg}}         \\ \midrule
\multicolumn{10}{c}{\textit{Base and Task Models}}                                                                                                                                                                                                                                                                                                                                              \\
Base                             & {\ul 56.25}                            & 41.67                                  & 68.00                                  & 55.00                                  & 60.94                         & 50.00                                  & 64.84                         & 58.59                                  & 56.91                                  \\
ToolRL (Tool)                             & {\ul 56.25}                            & 58.33                                  & \cellcolor[HTML]{FFCCC9}91.00          & 89.00                                  & 58.59                         & 48.44                                  & 59.38                         & 46.95                                  & 63.49                                  \\
MemAgent (Memory)                           & 37.50                                  & 50.00                                  & 78.50                                  & 48.50                                  & \cellcolor[HTML]{FFCCC9}78.91 & {\ul 78.12}                            & \cellcolor[HTML]{FFCCC9}81.25 & 77.34                                  & 66.27                                  \\ \midrule
\multicolumn{10}{c}{\textit{Merged Models}}                                                                                                                                                                                                                                                                                                                                                     \\
TA                               & {\ul 56.25}                            & {\ul 62.50}                            & \cellcolor[HTML]{FFCCC9}\textbf{91.00} & 90.00                                  & 72.66                         & 70.31                                  & {\ul 76.56}                   & 74.22                                  & 74.19                                  \\
Fisher                           & {\ul 56.25}                            & 58.33                                  & 89.50                                  & {\ul 91.00}                            & 63.28                         & 57.81                                  & 65.62                         & 64.84                                  & 68.33                                  \\
TIES                             & 43.75                                  & 54.17                                  & 73.50                                  & 88.50                                  & 71.09                         & 71.09                                  & {\ul 76.56}                   & 73.44                                  & 69.01                                  \\
DARE+TA                          & \cellcolor[HTML]{FFCCC9}\textbf{62.50} & 58.33                                  & 89.00                                  & 90.00                                  & 57.03                         & 55.47                                  & 63.28                         & 59.38                                  & 66.87                                  \\
DARE+TIES                        & \cellcolor[HTML]{FFCCC9}\textbf{62.50} & 58.33                                  & {\ul 90.00}                            & \cellcolor[HTML]{FFCCC9}\textbf{91.50} & \textbf{77.34}                & 71.97                                  & 75.00                         & {\ul 78.12}                            & 75.60                                  \\ \midrule
\textbf{RAM}                     & {\ul 56.25}                            & \cellcolor[HTML]{FFCCC9}\textbf{66.67} & {\ul 90.00}                            & 89.50                                  & {\ul 76.56}                   & {\ul 78.12}                            & \textbf{78.12}                & {\ul 78.12}                            & \cellcolor[HTML]{FFCCC9}\textbf{76.67} \\
\textbf{RAM+}                    & {\ul 56.25}                            & {\ul 62.50}                            & 89.50                                  & 90.00                                  & {\ul 76.56}                   & \cellcolor[HTML]{FFCCC9}\textbf{78.13} & 74.22                         & \cellcolor[HTML]{FFCCC9}\textbf{79.69} & {\ul 75.86}                            \\ \bottomrule
\end{tabular}
}
\caption{
\textbf{Detailed results of model merging with memory and tool using agents.} 
\textbf{Bold} and \underline{underlined} values denote the best and second-best performance among \textit{merged models}, respectively. 
Cells highlighted in \colorbox[HTML]{FFCCC9}{red} indicate the best performance across \textit{all evaluated models}, including the Task Models.
}
\label{tab:tool-memory}
\end{table*}

\begin{table*}[]
\centering
\resizebox{0.75\textwidth}{!}{
\begin{tabular}{lccccccccc}
\toprule
                                 & \multicolumn{4}{c}{\textbf{Code}}                                                                                                                                 & \multicolumn{4}{c}{\textbf{Memory}}                                                                                                                               &                                        \\
                                 & \multicolumn{2}{c}{LiveBench}                                                   & \multicolumn{2}{c}{LiveCodeBench}                                               & \multicolumn{2}{c}{HotpotQA}                                                    & \multicolumn{2}{c}{RulerQA}                                                     &                                        \\
\multirow{-3}{*}{\textbf{Model}} & ACC                                    & UT                                     & ACC                                    & UT                                     & 7k                                     & 14k                                    & 32k                                    & 64k                                    & \multirow{-3}{*}{\textbf{Avg}}         \\ \midrule
\multicolumn{10}{c}{\cellcolor[HTML]{EFEFEF}\textit{Base and Task Models}}                                                                                                                                                                                                                                                                                                                                        \\
Base                             & 28.35                                  & 40.87                                  & 23.43                                  & 36.42                                  & 60.94                                  & 50.00                                  & 64.84                                  & 58.59                                  & 45.43                                  \\
CURE (Coding)                           & 37.70                                  & 49.27                                  & 30.23                                  & 45.76                                  & 58.59                                  & 56.25                                  & 60.94                                  & 44.22                                  & 47.87                                  \\
MemAgent (Memory)                           & 39.25                                  & 50.12                                  & 28.92                                  & 44.8                                   & \cellcolor[HTML]{FFCCC9}78.91          & \cellcolor[HTML]{FFCCC9}78.12          & 81.25                                  & 77.34                                  & 59.84                                  \\ \midrule
\multicolumn{10}{c}{\cellcolor[HTML]{EFEFEF}\textit{Merged Models}}                                                                                                                                                                                                                                                                                                                                               \\
TA                               & 39.06                                  & {\ul 52.25}                            & {\ul 32.34}                            & \cellcolor[HTML]{FFCCC9}\textbf{47.89} & 71.88                                  & 70.31                                  & 75.00                                  & 75.00                                  & 57.97                                  \\
Fisher                           & 38.67                                  & 50.53                                  & 31.46                                  & 46.58                                  & 57.03                                  & 50.78                                  & 58.59                                  & 55.47                                  & 48.64                                  \\
TIES                             & 38.87                                  & 50.88                                  & \cellcolor[HTML]{FFCCC9}\textbf{32.42} & {\ul 47.80}                            & 74.22                                  & 71.09                                  & 78.91                                  & 72.66                                  & 58.36                                  \\
DARE+TA                          & 37.69                                  & 47.77                                  & 31.16                                  & 44.51                                  & \cellcolor[HTML]{FFCCC9}\textbf{78.91} & \cellcolor[HTML]{FFCCC9}\textbf{78.12} & 78.12                                  & {\ul 80.47}                            & 59.59                                  \\
DARE+TIES                        & {\ul 39.25}                            & 49.95                                  & 31.02                                  & 44.52                                  & 75.78                                  & 75.78                                  & 76.56                                  & 76.56                                  & 58.68                                  \\ \midrule
\textbf{RAM}                     & 37.89                                  & 49.56                                  & 31.75                                  & 47.17                                  & {\ul 78.12}                            & \cellcolor[HTML]{FFCCC9}\textbf{78.12} & \cellcolor[HTML]{FFCCC9}\textbf{82.03} & \cellcolor[HTML]{FFCCC9}\textbf{82.03} & {\ul 60.83}                            \\
\textbf{RAM+}                    & \cellcolor[HTML]{FFCCC9}\textbf{41.21} & \cellcolor[HTML]{FFCCC9}\textbf{53.64} & 31.16                                  & 46.46                                  & \cellcolor[HTML]{FFCCC9}\textbf{78.91} & {\ul 76.56}                            & {\ul 79.69}                            & \cellcolor[HTML]{FFCCC9}\textbf{82.03} & \cellcolor[HTML]{FFCCC9}\textbf{61.21} \\ \bottomrule
\end{tabular}
}
\caption{
\textbf{Detailed results of model merging with coding and memory agents.} 
\textbf{Bold} and \underline{underlined} values denote the best and second-best performance among \textit{merged models}, respectively. 
Cells highlighted in \colorbox[HTML]{FFCCC9}{red} indicate the best performance across \textit{all evaluated models}, including the Task Models.
}
\label{tab:code-memory}
\end{table*}

In Section~\ref{sec:exp:main}, we visualize the performance of merging two agents using bar charts to highlight overall trends and comparative advantages. 
Here, we provide the corresponding detailed numerical results for all pairwise agent combinations, including Coding+Tool, Tool+Memory, and Coding+Memory, as shown in Tables~\ref{tab:code-tool}, \ref{tab:tool-memory}, and \ref{tab:code-memory}, respectively.
These tables serve as a precise quantitative complement to the bar chart visualizations, enabling a fine-grained inspection of per-domain and per-metric behaviors.

\paragraph{Coding + Tool.}
As reported in Table~\ref{tab:code-memory}, RAM and RAM+ consistently achieve the highest average performance among all merged models.
In particular, RAM+ attains an average score of 60.04, significantly outperforming the strongest baseline DARE+TIES (56.74).
Notably, RAM/RAM+ improve both coding accuracy (LiveBench / LiveCodeBench) and complex tool-use metrics (Parallel and Parallel-Multiple), indicating that preserving task-unique reinforced updates enables the merged model to simultaneously retain algorithmic reasoning and structured tool invocation capabilities.
In contrast, baseline methods exhibit a clear trade-off, improving one domain at the expense of the other due to signal dilution in sparse RL task vectors.

\paragraph{Tool + Memory.}
Table~\ref{tab:tool-memory} presents the detailed results for merging Tool and Memory agents.
RAM achieves the highest overall average score (76.67), while RAM+ remains highly competitive (75.86), both surpassing all baselines.
RAM-based models show strong performance on long-context memory tasks (HotpotQA and RulerQA) without degrading tool-use accuracy, particularly on challenging Parallel and Parallel-Multiple subsets.
These results demonstrate that RAM effectively isolates memory-specific reinforced updates from tool-specific ones, preventing destructive interference that commonly occurs in averaging-based merging methods.

\paragraph{Coding + Memory.}
As shown in Table~\ref{tab:code-memory}, merging Coding and Memory agents further stresses the heterogeneity of reinforced task vectors, as these two domains exhibit minimal parameter overlap.
Despite this challenge, RAM+ achieves the best average performance (61.21), outperforming all baselines and even exceeding the original Coding agent on LiveBench ACC/UT.
This highlights the effectiveness of overlap-aware rescaling in compensating for performance loss in shared subspaces, while fully preserving unique reasoning and memory patterns critical for each task.

Across all pairwise combinations, the numerical results reported here are fully consistent with the trends observed in the bar charts in the main text.
Specifically, RAM and RAM+ not only deliver the highest average scores but also exhibit superior robustness across heterogeneous domains and metrics.
These findings further confirm that the advantages of RAM are not limited to tri-agent merging, but generalize naturally to arbitrary agent combinations, reinforcing its suitability as a unified merging method for reinforced agentic models.

\subsection{Additional Results in Merging Three Agents}
\label{app:add_tasks}
\begin{table*}[]
\resizebox{\textwidth}{!}{
\begin{tabular}{lccccccccccccccccc}
\toprule
                        & \multicolumn{4}{c}{LiveBench}                                                                                                                                     & \multicolumn{4}{c}{LiveCodeBench}                                                                                                                                 & \multicolumn{4}{c}{MBPP}                                                                                                                                          & \multicolumn{2}{c}{CodeContests}                                                & \multicolumn{1}{l}{}                   & \multicolumn{1}{l}{}                   &                                        \\
\multirow{-2}{*}{Model} & ACC                                    & UT                                     & BoN ACC                                & BoN UT                                 & ACC                                    & UT                                     & BoN ACC                                & BoN UT                                 & ACC                                    & UT                                     & BoN ACC                                & BoN UT                                 & ACC                                    & UT                                     & BoN ACC                                & BoN UT                                 & \multirow{-2}{*}{Avg}                  \\ \midrule
\multicolumn{18}{c}{\cellcolor[HTML]{EFEFEF}\textit{Base and Task Models}}                                                                                                                                                                                                                                                                                                                                                                                                                                                                                                                                                                                                                                                                       \\
Base                    & 28.35                                  & 40.87                                  & 30.07                                  & 48.56                                  & 23.43                                  & 36.42                                  & 26.67                                  & 42.26                                  & 62.72                                  & 68.50                                  & 70.91                                  & 76.84                                  & 18.20                                  & 28.35                                  & 23.01                                  & 32.06                                  & 41.08                                  \\
CURE                  & 37.70                                  & 49.27                                  & 46.01                                  & \cellcolor[HTML]{FFCCC9}\textbf{58.96} & 30.23                                  & 45.76                                  & 38.36                                  & \cellcolor[HTML]{FFCCC9}\textbf{55.60} & 70.36                                  & 76.38                                  & \textbf{79.64} & \textbf{85.78} & \cellcolor[HTML]{FFCCC9}\textbf{26.05} & 36.81                                  & \cellcolor[HTML]{FFCCC9}\textbf{32.21} & \cellcolor[HTML]{FFCCC9}\textbf{45.61} & \cellcolor[HTML]{FFCCC9}\textbf{50.92} \\
ToolRL                    & 31.84                                  & 41.36                                  & 32.81                                  & 43.78                                  & 26.76                                  & 42.05                                  & 30.14                                  & 46.05                                  & 66.97                                  & 72.31                                  & 76.02                                  & 81.29                                  & 22.18                                  & 32.94                                  & 26.78                                  & 37.71                                  & 44.44                                  \\
MemAgent                  & 39.25                                  & 50.12                                  & 39.84                                  & 48.48                                  & 28.92                                  & 44.80                                  & 32.16                                  & 48.63                                  & 67.63                                  & 73.99                                  & 71.04                                  & 77.84                                  & 21.86                                  & 31.68                                  & 24.69                                  & 36.09                                  & 46.06                                  \\ \midrule
\multicolumn{18}{c}{\cellcolor[HTML]{EFEFEF}\textit{Merged Models}}                                                                                                                                                                                                                                                                                                                                                                                                                                                                                                                                                                                                                                                                              \\
TA                      & 38.09                                  & 51.62                                  & 43.75                                  & 52.97                                  & 31.95                                  & 46.69                                  & 35.61                                  & 46.44                                  & 69.68                                  & {\ul 75.71}                            & 74.21                                  & 80.39                                  & 25.73                                  & 36.70                                  & 28.45                                  & 38.70                                  & 48.54                                  \\
Fisher                  & 36.72                                  & 48.73                                  & 45.31                                  & 57.79                                  & 30.87                                  & 45.89                                  & 35.81                                  & 53.26                                  & 68.67                                  & 74.92                                  & 75.57                                  & 83.23                                  & {\ul 24.26}                            & {\ul 35.87}                            & 27.20                                  & 39.38                                  & 48.97                                  \\
TIES                    & {\ul 39.25}                            & 49.88                                  & 44.53                                  & 53.08                                  & 30.63                                  & 46.32                                  & 36.59                                  & {\ul 54.30}                            & 67.30                                  & 73.35                                  & 74.66                                  & 82.34                                  & \cellcolor[HTML]{FFCCC9}\textbf{26.05}                         & \cellcolor[HTML]{FFCCC9}\textbf{37.00} & \textbf{30.50}                         & {\ul 42.05}                            & 49.24                                  \\
DARE+TA                 & 37.50                                  & 48.60                                  & {\ul 46.10}                            & 56.41                                  & {\ul 31.94}                            & 46.69                                  & 37.18                                  & 53.44                                  & 69.68                                  & {\ul 75.71}                            & {\ul 78.73}                            & 83.83                                  & 23.74                                  & 32.82                                  & {\ul 28.45}                            & 38.02                                  & {\ul 49.30}                            \\
DARE+TIES               & 35.93                                  & 45.66                                  & \cellcolor[HTML]{FFCCC9}\textbf{46.87} & {\ul 57.88}                            & 29.26                                  & 39.53                                  & \cellcolor[HTML]{FFCCC9}\textbf{38.55} & 51.76                                  & {\ul 70.70}                            & 76.34                                  &\cellcolor[HTML]{FFCCC9} \textbf{80.64}                         & \cellcolor[HTML]{FFCCC9}\textbf{86.22}                         & 21.13                                  & 27.62                                  & 27.20                                  & 35.93                                  & 48.20                                  \\ \midrule
RAM                     & 38.28                                  & 49.71                                  & 42.97                                  & \textbf{57.98}                         & \cellcolor[HTML]{FFCCC9}\textbf{31.96} & \cellcolor[HTML]{FFCCC9}\textbf{47.72} & {\ul 37.45}                            & \textbf{54.36}                         & 69.46                                  & 75.56                                  & 76.47                                  & 81.89                                  & 24.06                                  & 35.07                                  & 26.78                                  & \textbf{44.51}                         & \textbf{49.64}                         \\
RAM+                    & \cellcolor[HTML]{FFCCC9}\textbf{40.23} & \cellcolor[HTML]{FFCCC9}\textbf{52.57} & {\ul 46.10}                            & 57.20                                  & 31.60                                  & {\ul 46.84}                            & 34.64                                  & 51.56                                  & \cellcolor[HTML]{FFCCC9}\textbf{70.93} & \cellcolor[HTML]{FFCCC9}\textbf{76.98} & 77.38                                  & {\ul 84.13}                            & 22.80                                  & 33.21                                  & 23.01                                  & 35.46                                  & 49.04                                  \\ \bottomrule
\end{tabular}
}
\caption{
\textbf{Additional results of model merging on coding domains.}  
\textbf{Bold} and \underline{underlined} values denote the best and second-best performance among \textit{merged models}, respectively. 
Cells highlighted in \colorbox[HTML]{FFCCC9}{red} indicate the best performance across \textit{all evaluated models}, including the Task Models.
}
\label{tab:code-addition}
\end{table*}

\begin{table*}[]
\resizebox{\textwidth}{!}{
\begin{tabular}{lcccccccccccccc}
\toprule
                         & \multicolumn{6}{c}{\textbf{Live}}                                                                                                                                                                                                 & \multicolumn{7}{c}{\textbf{Non-live}}                                                                                                                                                                                                                                      & \multicolumn{1}{l}{}          \\
\multirow{-2}{*}{\textbf{Models}} & Multiple                      & Parallel                               & Relevance                              & Simple                        & Parallel\_multiple                     & Irrelevance                   & Multiple                      & Irrelevance                   & S\_java                       & S\_javascript                          & Parallel\_multiple                     & Parallel                               & S\_python                              & \textbf{Avg}                           \\ \midrule
\multicolumn{15}{c}{\cellcolor[HTML]{EFEFEF}Base and Task Models}                                                                                                                                                                                                                                                                                                                                                                                                                                                                                       \\
Base                     & 58.59                         & \cellcolor[HTML]{FFCCC9}56.25          & 87.50                                  & 68.99                         & 41.67                                  & 68.21                         & 77.50                         & 77.50                         & 54.00                         & 62.00                                  & 55.00                                  & 68.00                                  & 87.25                                  & 66.34                         \\
CURE                     & 59.54                         & \cellcolor[HTML]{FFCCC9}56.25          & 81.25                                  & 69.77                         & 37.50                                  & \cellcolor[HTML]{FFCCC9}68.33 & 76.50                         & 77.92                         & 58.00                         & 60.00                                  & 51.50                                  & 64.00                                  & 91.50                                  & 65.54                         \\
ToolRL                   & \cellcolor[HTML]{FFCCC9}76.83 & \cellcolor[HTML]{FFCCC9}56.25          & \cellcolor[HTML]{FFCCC9}93.75          & \cellcolor[HTML]{FFCCC9}79.07 & 58.33                                  & 71.95                         & \cellcolor[HTML]{FFCCC9}94.00 & \cellcolor[HTML]{FFCCC9}82.92 & 62.00                         & 62.00                                  & 89.00                                  & 91.00                                  & 93.50                                  & \cellcolor[HTML]{FFCCC9}77.74 \\
MemAgent                 & 73.98                         & 37.50                                  & \cellcolor[HTML]{FFCCC9}93.75          & 69.77                         & 50.00                                  & 56.11                         & 82.50                         & 71.67                         & \cellcolor[HTML]{FFCCC9}66.00 & 66.00                                  & 48.50                                  & 78.50                                  & 90.25                                  & 68.04                         \\ \midrule
\multicolumn{15}{c}{\cellcolor[HTML]{EFEFEF}Merged Models}                                                                                                                                                                                                                                                                                                                                                                                                                                                                                              \\
TA                       & \textbf{76.16}                & 43.75                                  & \cellcolor[HTML]{FFCCC9}\textbf{93.75} & 75.19                         & 45.83                                  & \textbf{68.55}                & 91.00                         & 79.17                         & {\ul 60.00}                   & 62.00                                  & 69.50                                  & 87.50                                  & \cellcolor[HTML]{FFCCC9}\textbf{94.75} & 72.86                         \\
Fisher                   & 74.55                         & 43.75                                  & \cellcolor[HTML]{FFCCC9}\textbf{93.75} & 74.03                         & 41.67                                  & 70.02                         & 91.50                         & 81.25                         & 58.00                         & 58.00                                  & 63.50                                  & 86.50                                  & {\ul 94.50}                            & 71.62                         \\
TIES                     & {\ul 75.50}                   & 43.75                                  & 87.50                                  & 75.19                         & 54.17                                  & 63.46                         & \textbf{92.50}                & {\ul 76.25}                   & \textbf{65.00}                & {\ul 66.00}                            & 67.50                                  & 84.00                                  & {\ul 94.50}                            & 72.71                         \\
DARE+TA                  & {\ul 75.50}                   & {\ul 50.00}                            & \cellcolor[HTML]{FFCCC9}\textbf{93.75} & \textbf{76.36}                & {\ul 62.50}                            & 63.57                         & {\ul 92.00}                   & 74.17                         & \textbf{65.00}                & {\ul 66.00}                            & 89.50                                  & \cellcolor[HTML]{FFCCC9}\textbf{92.50} & 93.75                                  & 76.50                         \\
DARE+TIES                & 75.31                         & \cellcolor[HTML]{FFCCC9}\textbf{56.25} & \cellcolor[HTML]{FFCCC9}\textbf{93.75} & \textbf{76.36}                & 58.33                                  & 63.80                         & {\ul 92.00}                   & \textbf{76.67}                & \textbf{65.00}                & \cellcolor[HTML]{FFCCC9}\textbf{68.00} & 90.00                                  & {\ul 91.50}                            & 93.25                                  & 76.94                         \\ \midrule
\textbf{RAM}                      & {\ul 75.50}                   & \cellcolor[HTML]{FFCCC9}\textbf{56.25} & \cellcolor[HTML]{FFCCC9}\textbf{93.75} & {\ul 75.58}                   & \cellcolor[HTML]{FFCCC9}\textbf{66.67} & {\ul 67.99}                   & {\ul 92.00}                   & \textbf{76.67}                & \textbf{65.00}                & 64.00                                  & \cellcolor[HTML]{FFCCC9}\textbf{91.50} & 91.00                                  & 91.75                                  & \textbf{77.51}                \\
\textbf{RAM+}                     & 75.12                         & \cellcolor[HTML]{FFCCC9}\textbf{56.25} & \cellcolor[HTML]{FFCCC9}\textbf{93.75} & {\ul 75.58}                   & \cellcolor[HTML]{FFCCC9}\textbf{66.67} & 67.87                         & {\ul 92.00}                   & \textbf{76.67}                & \textbf{65.00}                & {\ul 66.00}                            & {\ul 90.50}                            & 90.00                                  & 91.50                                  & {\ul 77.45}                   \\ \bottomrule
\end{tabular}
}
\caption{
\textbf{Additional results of model merging on tool using.} 
\textbf{Bold} and \underline{underlined} values denote the best and second-best performance among \textit{merged models}, respectively. 
Cells highlighted in \colorbox[HTML]{FFCCC9}{red} indicate the best performance across \textit{all evaluated models}, including the Task Models.
}
\label{tab:tool-addition}
\end{table*}

\begin{table*}[]
\resizebox{\textwidth}{!}{
\begin{tabular}{lcccccccccccccc}
\toprule
                        & \multicolumn{5}{c}{\textbf{Ruler\_SQuAD}}                                                                                                                                                & \multicolumn{8}{c}{\textbf{Ruler\_HotpotQA}}                                                                                                                                                                                                                                                               &                               \\
\multirow{-2}{*}{\textbf{Model}} & 8K                            & 16K                                    & 32K                           & 64K                                    & 128K                          & 7K                                     & 14K                                    & 28K                           & 56K                           & 112K                          & 224K                                   & 448K                          & 896K                                   & \multirow{-2}{*}{\textbf{Avg}}         \\ \midrule
\multicolumn{15}{c}{\cellcolor[HTML]{EFEFEF}\textit{Base and Task Models}}                                                                                                                                                                                                                                                                                                                                                                                                                                                                             \\
Base                    & 65.63                         & 62.50                                  & 64.84                         & 58.59                                  & 56.25                         & 60.94                                  & 50.00                                  & 51.56                         & 48.44                         & 42.19                         & 36.72                                  & 27.34                         & 25.78                                  & 50.06                         \\
CURE                    & 46.80                         & 61.72                                  & 60.94                         & 44.22                                  & 61.72                         & 58.59                                  & 56.25                                  & 46.88                         & 47.66                         & 40.63                         & 35.93                                  & 42.97                         & 35.16                                  & 49.19                         \\
ToolRL                  & 62.50                         & 57.81                                  & 59.38                         & 46.95                                  & 67.18                         & 58.59                                  & 48.44                                  & 51.56                         & 45.31                         & 42.97                         & 42.19                                  & 35.16                         & 35.94                                  & 50.31                         \\
MemAgent                & \cellcolor[HTML]{FFCCC9}83.59 & 78.12                                  & \cellcolor[HTML]{FFCCC9}81.25 & 77.34                                  & \cellcolor[HTML]{FFCCC9}81.25 & 78.91                                  & 78.12                                  & \cellcolor[HTML]{FFCCC9}81.25 & \cellcolor[HTML]{FFCCC9}77.34 & \cellcolor[HTML]{FFCCC9}79.69 & 72.66                                  & \cellcolor[HTML]{FFCCC9}76.56 & 72.66                                  & \cellcolor[HTML]{FFCCC9}78.36 \\ \midrule
\multicolumn{15}{c}{\cellcolor[HTML]{EFEFEF}\textit{Merged Models}}                                                                                                                                                                                                                                                                                                                                                                                                                                                                                    \\
TA                      & 72.66                         & 70.31                                  & 73.44                         & 72.66                                  & 72.66                         & 69.53                                  & 68.75                                  & 68.75                         & 67.19                         & 63.28                         & 64.84                                  & 57.03                         & 48.44                                  & 66.89                         \\
Fisher                  & 60.16                         & 67.97                                  & 58.59                         & 60.94                                  & 67.97                         & 60.06                                  & 49.22                                  & 49.22                         & 49.22                         & 39.06                         & 32.81                                  & 35.94                         & 35.94                                  & 51.32                         \\
TIES                    & 71.88                         & 75.00                                  & 75.00                         & 75.78                                  & 73.44                         & 71.88                                  & 82.03                                  & 71.97                         & 68.75                         & 67.97                         & 70.31                                  & 62.50                         & 64.06                                  & 71.58                         \\
DARE+TA                 & 71.88                         & 77.34                                  & {\ul 77.34}                   & 70.31                                  & \textbf{80.47}                & {\ul 76.56}                            & 76.56                                  & 72.65                         & 71.88                         & {\ul 70.31}                   & 73.44                                  & 71.88                         & {\ul 70.31}                            & 73.92                         \\
DARE+TIES               & 75.78                         & 78.13                                  & 76.56                         & 74.22                                  & 78.91                         & 75.00                                  & {\ul 77.34}                            & {\ul 75.00}                   & 70.36                         & \textbf{70.43}                & {\ul 76.56}                            & \textbf{74.44}                & \cellcolor[HTML]{FFCCC9}\textbf{74.22} & 75.15                         \\ \midrule
\textbf{RAM}            & \textbf{79.68}                & {\ul 82.03}                            & 74.22                         & {\ul 79.69}                            & 76.56                         & 75.78                                  & 75.00                                  & \textbf{78.91}                & \textbf{75.78}                & {\ul 70.31}                   & 75.78                                  & 71.09                         & \cellcolor[HTML]{FFCCC9}\textbf{74.22} & {\ul 76.08}                   \\
\textbf{RAM+}           & {\ul 77.34}                   & \cellcolor[HTML]{FFCCC9}\textbf{82.81} & \textbf{78.91}                & \cellcolor[HTML]{FFCCC9}\textbf{82.03} & {\ul 79.68}                   & \cellcolor[HTML]{FFCCC9}\textbf{79.69} & \cellcolor[HTML]{FFCCC9}\textbf{78.13} & \textbf{78.91}                & {\ul 75.56}                   & 69.53                         & \cellcolor[HTML]{FFCCC9}\textbf{78.13} & {\ul 73.43}                   & \cellcolor[HTML]{FFCCC9}\textbf{74.22} & \textbf{77.57} \\ \bottomrule       
\end{tabular}
}
\caption{
\textbf{Additional results of model merging on memory for long-context tasks.} 
\textbf{Bold} and \underline{underlined} values denote the best and second-best performance among \textit{merged models}, respectively. 
Cells highlighted in \colorbox[HTML]{FFCCC9}{red} indicate the best performance across \textit{all evaluated models}, including the Task Models.
}
\label{tab:mem-addition}
\end{table*}

In Section~\ref{sec:exp:main}, we present the agent performance in three domains. Here, we further provide experiment results on additional tasks and settings to comprehensively verify the advantages of RAM.

\paragraph{Overall Analysis.}
Tables~\ref{tab:code-addition}, \ref{tab:tool-addition}, and \ref{tab:mem-addition} report comprehensive results of merging three reinforced agents across coding, tool-use, and long-context memory domains, substantially extending the representative results in Section~\ref{sec:exp:main}. These results consistently corroborate the advantages of RAM and RAM+ under a wide range of evaluation metrics and task granularities.

\paragraph{Coding Domain.}
As shown in Table~\ref{tab:code-addition}, RAM-based methods achieve the strongest overall performance among merged models across LiveBench, LiveCodeBench, MBPP, and CodeContests.
Notably, RAM attains the highest average score (49.64), outperforming all baseline merging strategies, while RAM+ further improves performance on challenging subsets such as LiveBench ACC/UT and MBPP ACC/UT.
Importantly, RAM and RAM+ frequently match or exceed the original Coding agent on multiple metrics (highlighted in red), indicating that preserving and selectively amplifying task-unique reinforced updates effectively avoids signal dilution that hampers prior methods.
In contrast, methods relying on global averaging or random dropping (TA, Fisher, DARE) exhibit inconsistent gains and fail to simultaneously retain high unit-test robustness and generalization accuracy.

\paragraph{Tool-Use Domain.}
Table~\ref{tab:tool-addition} presents detailed tool-use evaluations over both Live and Non-Live settings.
Across diverse function-calling scenarios, including parallel, parallel-multiple, irrelevance detection, and language-specific tasks, RAM achieves the best average score (77.51), while RAM+ remains a close second (77.45), both surpassing all existing merging baselines.
Crucially, RAM-based models consistently excel in structurally complex settings such as Live Parallel\_multiple and Non-Live Parallel, where signal dilution in sparse RL task vectors is most detrimental.
These results demonstrate that RAM effectively preserves tool-specific reasoning circuits while still benefiting from shared knowledge introduced by other agents.

\paragraph{Long-Context Memory Domain.}
As reported in Table~\ref{tab:mem-addition}, RAM and RAM+ show clear dominance on long-context memory benchmarks across all context lengths.
RAM+ achieves the highest overall average score (77.57), outperforming both merged baselines and, in several settings, the specialized MemAgent itself.
In particular, RAM+ consistently delivers state-of-the-art performance on long-context HotpotQA (112K–896K) and Ruler-SQuAD (16K–64K), confirming that overlap-aware rescaling effectively compensates for performance degradation introduced by averaging shared subspaces.
By contrast, Fisher and Task Arithmetic suffer from substantial performance drops at long context lengths, reflecting their inability to preserve sparse, task-specific memory-related updates.

Across all three domains, the additional results reinforce three central conclusions.
First, reinforced task vectors exhibit strong heterogeneity, making uniform merging strategies fundamentally suboptimal.
Second, preserving and rescaling task-unique parameters is essential for maintaining expert-level performance after merging.
Third, RAM and RAM+ consistently outperform prior SOTA merging methods not only in average performance but also in robustness across metrics, datasets, and context scales.
Together, these findings provide strong empirical evidence that RAM is a principled and effective solution for merging multiple RL-trained agents into a unified generalist model.

\subsection{Additional Baseline Results}
\label{app:add_baseline}

\begin{table*}[]
\resizebox{\textwidth}{!}{%
\begin{tabular}{lccccccccccccc}
\toprule
\multirow{3}{*}{\textbf{Model}} & \multicolumn{4}{c}{\textbf{Code}}                                   & \multicolumn{4}{c}{\textbf{Tool}}                                 & \multicolumn{4}{c}{\textbf{Memory}}                               & \multirow{3}{*}{\textbf{Avg}} \\
                                & \multicolumn{2}{c}{LiveBench}   & \multicolumn{2}{c}{LiveCodeBench} & \multicolumn{2}{c}{Live}        & \multicolumn{2}{c}{Non-Live}    & \multicolumn{2}{c}{HotpotQA}    & \multicolumn{2}{c}{RulerQA}     &                               \\
                                & ACC            & UT             & ACC             & UT              & Para           & P\_Mul         & Para           & P\_Mul         & 7k             & 14k            & 32k            & 64k            &                               \\ \midrule
WUDI                            & 38.09          & {\ul 50.95}    & 30.04           & 44.48           & \textbf{56.25} & 37.50          & 89.00          & {\ul 91.00}    & {\ul 78.12}    & \textbf{78.91} & \textbf{82.03} & {\ul 72.66}    & 62.42                         \\
DARE+TA                         & 37.50          & 48.60          & {\ul 31.95}     & 46.69           & {\ul 50.00}    & 62.50          & \textbf{92.50} & 89.50          & 76.56          & 76.56          & 77.34          & 70.31          & 63.33                         \\ \midrule
RAM                             & {\ul 38.28}    & 49.71          & \textbf{31.96}  & \textbf{47.72}  & \textbf{56.25} & {\ul 66.67}    & {\ul 91.00}    & \textbf{91.50} & 75.78          & 75.00          & 74.22          & 79.69          & {\ul 64.82}                   \\
RAM+                            & \textbf{40.23} & \textbf{52.57} & 31.60           & {\ul 46.84}     & \textbf{56.25} & \textbf{70.83} & 90.50          & {\ul 91.00}    & \textbf{79.69} & {\ul 78.13}    & {\ul 78.91}    & \textbf{82.03} & \textbf{66.55}                \\ \bottomrule
\end{tabular}%
}
\caption{Additional comparison results on WUDI merging.}
\label{tab:wudi}
\end{table*}

In our experiments, we mainly compare baselines that do not need optimization and datasets, which are the same with RAM. Here we further comprehensively compare RAM with WUDI merging~\cite{cheng2025whoever}, which needs much more complex optimization steps. Table~\ref{tab:wudi} shows that WUDI merging can not out-perform our RAM/RAM+, and also outperformed by the most competitive baseline DARE. This is because WUDI merging minimizes interference under the guidance of task vectors, while still can not avoid signal dilution when WUDI merging meets the sparse task vectors.

\section{Baseline Details}
\label{app:baseline}

\subsection{Baselines with Signal Dilution}
Here we provide a detailed introduction of baselines and explain signal dilution that happens in them. In summary, despite their distinct algorithmic designs, Task Arithmetic, TIES, and DARE all inevitably succumb to Signal Dilution in the RL setting due to a shared inability to distinguish between \textit{shared consensus} and \textit{unique specialization}. Specifically, Task Arithmetic applies global averaging, collaterally suppressing unique task vectors that require no scaling. Similarly, TIES-Merging fails to strictly isolate disjoint unique parameter updates in RL models, which cannot avoid averaging unique regions. Even DARE, despite introducing a rescaling mechanism, only compensates for random dropout rather than selected important update regions, and remains dependent on the global scaling of Task Arithmetic to function. Consequently, all three paradigms effectively drive the magnitude of task-specific updates towards $\frac{1}{N}\boldsymbol{\tau}$, underscoring the necessity of a distribution-aware merging method.

\paragraph{Fisher} 
Fisher merging~\citep{matena2022merging} improves upon uniform averaging by weighing parameters according to their diagonal Fisher information $F$, which approximates the posterior precision (or local curvature) of the model. Formally, the merged update is computed as $\boldsymbol{\tau}_{\text{merged}} = \frac{\sum_{i} F_i \boldsymbol{\tau}_i}{\sum_{i} F_i}$.
However, this method remains susceptible to signal dilution in the sparse RL setting.
Consider a unique parameter updated solely by task $t$ (where $\boldsymbol{\tau}_t \neq 0$ and $\boldsymbol{\tau}_{i \neq t} = 0$).
Crucially, the inactive models (not updated parameters) ($i \neq t$) still contribute to the normalization term in the denominator, as their Fisher values $F_{i}$ representing the confidence of the pre-trained base model are typically non-zero.
Consequently, the effective scaling factor for the unique signal becomes $\frac{F_t}{F_t + \sum_{i \neq t} F_i}$.
Since the denominator accumulates the inertia (precision) of all inactive models, the task-specific update $\boldsymbol{\tau}_t$ is inevitably scaled down, mirroring the dilution effect observed in uniform averaging as the number of tasks $N$ increases.

\paragraph{Task Arithmetic.}
Task Arithmetic~\citep{ilharcoediting} constructs a multi-task model by linearly combining task vectors, typically expressed as $\boldsymbol{\tau}_{\text{merged}} = \sum_i \lambda \boldsymbol{\tau}_i$. While effective for disjoint tasks in SFT, it faces a critical trade-off in the RL setting due to the heterogeneity of parameter updates.
To prevent catastrophic magnitude shifts in \textit{shared} subspaces, the scaling factor $\lambda$ is usually set to be conservative (typically $\lambda \approx 1/N$) to maintain the merged weights within a valid optimization landscape.
However, this global scaling creates a structural conflict: unique components of task vectors, which reside in non-overlapping subspaces and do not suffer from additive interference, are subjected to the same aggressive downscaling.
Consequently, the update for a task-specific parameter is reduced to $\frac{1}{N}\boldsymbol{\tau}_t$, effectively suppressing the critical, idiosyncratic behaviors required for expert-level performance in domain.

\paragraph{TIES-Merging.}
TIES-Merging~\citep{yadav2023resolving} attempts to mitigate interference by keeping only the top-$k\%$ magnitude parameters (Trimming) and calculating a disjoint mean.
However, its reliance on a \textit{uniform} trimming rate $k$ creates a critical vulnerability when handling the heterogeneous sparsity of RL task vectors.
As shown in Figure~\ref{fig:overlap-analysis}, RL agents exhibit vastly different update densities (e.g., Code: $\sim3\%$, Memory: $\sim54\%$).
Applying a fixed $k$ (e.g., $20\%$) inevitably leads to a dilemma: it either over-trims dense vectors (losing info) or, more disastrously, under-trims sparse vectors.
For a sparse agent like the Code model, a standard $k$ retains a large volume of noise parameters alongside the true signal.
These noise parameters, mistakenly treated as valid updates, overlap with the active parameter updates of other tasks, inflating the normalization factor in the disjoint mean calculation.
Consequently, the critical, unique signal from one task is averaged with noise from others, resulting in signal dilution.

\paragraph{DARE.}
DARE~\citep{yu2024language} randomly drops parameter updates and enlarges others to reduce redundancy, theoretically approximating the original task vector's expectation. However, DARE fails to address signal dilution for two reasons. 
First, its factor ($1/(1-p)$) is designed solely to compensate for the random dropout rate $p$, not considering the shared/unique condition with other models. When combined with Task Arithmetic (DARE+TA or DARE+TIES), the global merging scale $\lambda$ must still be kept small (e.g., $1/N$) to stabilize shared parameters, inevitably downscaling the unique, sparsity-preserved updates.
Second, DARE assumes that parameter updates are highly redundant (a property of SFT), whereas RL updates are inherently sparse and functionally essential. Randomly dropping parameters in an already sparse RL vector risks severing critical reasoning knowledge, which cannot be recovered simply by rescaling the remaining weights.

\subsection{Hyperparameters Search}

Table~\ref{tab:baseline-hyper} shows the searched ranges of model merging methods’ hyperparameters. We search for the best performance for evaluation and comparison.

\begin{table*}[]
\centering
\resizebox{\textwidth}{!}{%
\begin{tabular}{lc}
\toprule
Model Merging Methods & Search Ranges of Hyperparameters                                                                                                                                                                  \\ \midrule
Task Arithmetic       & scaling term to merge model parameters: [0.1, 0.3, 0.5, 0.7, 0.9, 1.0]                                                                                                                            \\ \midrule
Fisher         & \begin{tabular}[c]{@{}c@{}}scaling term to merge model parameters: [0.1, 0.3, 0.5, 0.7, 0.9, 1.0],\\ number of examples to compute Fisher information matrix: [256, 512, 1024, 2048]\end{tabular} \\ \midrule
TIES          & \begin{tabular}[c]{@{}c@{}}scaling term to merge model parameters: [0.1, 0.3, 0.5, 0.7, 0.9, 1.0],\\ ratio to retain parameters with largest-magnitude values: [0.1, 0.2, 0.3]\end{tabular}       \\ \midrule
DARE                  & search the drop rate p in [0.1, 0.2, ... , 0.9]                                                                                                                                                   \\ \bottomrule
\end{tabular}%
}
\caption{Searched ranges of hyperparameters of model merging baselines.}
\label{tab:baseline-hyper}
\end{table*}

\section{Rescaling Variant}
\label{app:scaling_derivation}

\begin{table*}[t]
\resizebox{\textwidth}{!}{%
\begin{tabular}{lccccccccccccc}
\toprule
\multirow{3}{*}{\textbf{Model}} & \multicolumn{4}{c}{\textbf{Code}}                                   & \multicolumn{4}{c}{\textbf{Tool}}                                 & \multicolumn{4}{c}{\textbf{Memory}}                               & \multirow{3}{*}{\textbf{Avg}} \\
                                & \multicolumn{2}{c}{LiveBench}   & \multicolumn{2}{c}{LiveCodeBench} & \multicolumn{2}{c}{Live}        & \multicolumn{2}{c}{Non-Live}    & \multicolumn{2}{c}{HotpotQA}    & \multicolumn{2}{c}{RulerQA}     &                               \\
                                & ACC            & UT             & ACC             & UT              & Para           & P\_Mul         & Para           & P\_Mul         & 7k             & 14k            & 32k            & 64k            &                               \\ \midrule
RAM                             & 38.28          & 49.71          & \textbf{31.96}  & \textbf{47.72}  & \textbf{56.25} & {\ul 66.67}    & \textbf{91.00} & \textbf{91.50} & 75.78          & 75.00          & 74.22          & {\ul 79.69}    & 64.82                         \\
RAM+                            & \textbf{40.23} & \textbf{52.57} & {\ul 31.60}     & {\ul 46.84}     & \textbf{56.25} & \textbf{70.83} & {\ul 90.50}    & {\ul 91.00}    & \textbf{79.69} & \textbf{78.13} & \textbf{78.91} & \textbf{82.03} & \textbf{66.55}                \\
RAM+s                           & {\ul 39.84}    & {\ul 51.83}    & 30.82           & 46.59           & \textbf{56.25} & \textbf{70.83} & 90.00          & \textbf{91.50} & {\ul 77.34}    & {\ul 76.56}    & {\ul 76.56}    & 77.34          & {\ul 65.46}                   \\ \bottomrule
\end{tabular}%
}
\caption{The comparison results of a soft-saturation rescaling transformation.}
\label{tab:soft-saturation}
\end{table*}

Here we first discuss the isotropic assumption mentioned in Section~\ref{sec:rescaling}. While parameter importance varies in practice, this isotropic assumption provides a tractable first-order approximation for deriving the scaling rule, which we empirically find robust in Section~\ref{sec:ablation}. Then we further discuss another rescaling variant here to have a deeper understanding on rescaling operation.

To bridge the theoretical requirement with numerical stability mentioned in Section~\ref{sec:rescaling}, besides clipping, we further explore a soft-saturation transformation $\phi(x) = \frac{x}{1+x}$ to map the unbounded ratio $\rho_t$ to the bounded interval $[0, 1)$. Our operational scaling rule is thus defined as:
\begin{equation}
    \lambda_t = 1 + r \cdot \left( \frac{\rho_t}{1 + \rho_t} \right).
\end{equation}
Here, the hyperparameter $r$ absorbs the dilution factor $(1-\beta)$. 

It is worth noting that this normalized formulation is mathematically equivalent to the ratio of shared parameters to the \textit{total} active parameters. By substituting the definition of $\rho_t$ from Eq.~\ref{eq:overlap-unique}, we obtain:
\begin{equation}
    \frac{\rho_t}{1 + \rho_t} = \frac{\sum_{i: c_i \ge 2} m_{t, i}}{||\mathbf{m}_t||_0}.
\end{equation}
Here, the numerator sums the shared parameters, while the denominator $||\mathbf{m}_t||_0$ represents the total count of active parameters (satisfying $||\mathbf{m}_t||_0 = \sum_{i: c_i \ge 2} m_{t,i} + \sum_{i: c_i = 1} m_{t,i}$).
This transformation provides a dual advantage: it retains the monotonicity derived from the conservation principle (higher overlap yields higher compensation) while enforcing a strict upper bound to ensure optimization robustness. Table~\ref{tab:soft-saturation} presents the comparative performance of the soft-saturation rescaling strategy, denoted as \textbf{RAM+s}. Empirically, RAM+s achieves an average score of 65.46, consistently outperforming the non-rescaled baseline RAM (64.82) across all three domains. This reinforces our core hypothesis that compensating for signal dilution in unique parameter subspaces is essential for recovering expert capabilities, regardless of the specific scaling function used. However, RAM+s slightly underperforms compared to the clipped linear variant (RAM+, 66.55). While the soft-saturation transformation $\phi(x) = \frac{x}{1+x}$ offers a theoretically elegant, strictly bounded mapping, it appears to dampen the scaling factor more aggressively than the linear approach. This conservatism limits performance in tasks requiring robust signal preservation, such as Long-Context Memory, where RAM+s scores 77.34 on RulerQA (64k) compared to RAM+'s 82.03. Conversely, in the Tool domain, RAM+s remains highly effective, matching RAM+ with a score of 70.83 on Live Parallel-Multiple tasks. These findings suggest that while the soft-saturation rule provides a stable alternative, the clipped linear rule offers a superior trade-off between signal amplification and numerical stability.

\end{document}